\documentclass[11pt]{article}

\usepackage[preprint]{acl}

\usepackage{times}
\usepackage{latexsym}

\usepackage[T1]{fontenc}

\usepackage[utf8]{inputenc}

\usepackage{microtype}
\usepackage{multirow}

\usepackage{inconsolata}

\usepackage{graphicx}

\usepackage{hyperref}       
\usepackage{url}            
\usepackage{booktabs}       
\usepackage{amsfonts, amsmath}       
\usepackage{nicefrac}       
\usepackage{microtype}      
\usepackage[table]{xcolor}         
\usepackage{times}  
\usepackage{helvet}  
\usepackage{courier}  
\usepackage{natbib}  
\usepackage[skins, breakable]{tcolorbox} 
\usepackage{caption} 
\usepackage{svg}
\usepackage{algorithm}
\usepackage{algorithmic}
\usepackage{array}
\usepackage{booktabs}
\usepackage{threeparttable}
\usepackage{newfloat}
\usepackage{listings}
\usepackage{xcolor}

\usepackage{alltt}
\tcbset{fontupper=\footnotesize}

\title{Belief Without Behavior: Measuring the Translation of Theory of Mind into Coordinated Social Action in Vision-Language Models}

\author{Tonglin Yan \and Gregoire Sergeant-Perthuis \and David Rudrauf \\
        CIAMS, Université Paris-Saclay \\ CQSB, Sorbonne Université \\ CIAMS, Université Paris-Saclay\\ \texttt{tonglin.yan@universite-paris-saclay.fr}}

\begin{document}

\maketitle

\begin{abstract}
Effective social interaction requires agents to translate mental state inferences into coordinated behavioral signals across verbal and nonverbal channels simultaneously. Yet existing benchmarks evaluate theory of mind (ToM) reasoning and embodied behavior in isolation, leaving unmeasured the gap between social inference and social action. We introduce MOSAIC (Multimodal Orchestration of Social Action, Inference, and Communication), a controlled benchmark in which two embodied agents interact across cooperative and competitive scenarios requiring integration of verbal statements, spatial trajectories, gaze direction, and facial expression under systematically varied ToM constraints. Evaluating 13 models, including 11 VLMs, across 200 trials per model, we find that VLMs fail to produce behaviors consistent with the expected outcomes under ToM-order constraints, and that imposing explicit ToM-order constraints produces no reliable behavioral change aligned with the specified reasoning level. Signal-level analysis reveals two sequential bottlenecks: most models cannot produce directionally coherent nonverbal signals, and even when signals are present, VLM agents fail to interpret others behaviors and react to them. PCM-LLM, included as a structured architectural reference point with an explicit ToM module, succeeds across all conditions, suggesting that explicit belief-action coupling is a sufficient ingredient for this class of tasks.\footnote{Code available at: \url{https://anonymous.4open.science/r/MOSAIC-50EE}}
\end{abstract}

\section{Introduction}


Theory of Mind (ToM), broadly defined as the ability to attribute mental states to oneself and others and to use those attributions to explain and predict behavior \citep{premackDoesChimpanzeeHave1978, byomTheoryMindMechanisms2013, wellmanChildsTheoryMind1992}, is widely regarded as a foundational component of human social cognition \citep{frithSocialBrain2007}. Yet mental state attribution is only part of what social interaction demands: translating belief inferences into coordinated behavioral responses across multiple channels simultaneously, including verbal utterances, spatial movement, gaze, and facial expression, is itself a distinct and non-trivial requirement \citep{notaFacialSignalsSocial2021, burgoonNonverbalBehaviorsSpeak2021}. We term this integrated capacity \emph{embodied social intelligence}, distinguishing it from the physical task completion that dominates existing embodied evaluation. Whereas physical agents act on the world to change its state, socially intelligent agents act on the minds of others to influence their beliefs and behavior. In skilled human social interaction, mental state attribution and behavioral response are tightly coupled: inferred beliefs about others reliably guide the selection of appropriately timed, appropriately channeled actions across verbal and nonverbal modalities \citep{christensenIanApperlyMindreaders2013, burgoonNonverbalBehaviorsSpeak2021, sebanzJointActionBodies2006a}. Whether current AI systems achieve the same integration, such that belief inferences about others translate into coordinated multimodal action, remains an open empirical question \citep{guSimpleToMExposingGap2024, houEgoSocialArenaBenchmarkingSocial2025} that existing evaluation frameworks are not designed to answer \citep{ullmanLargeLanguageModels2023, byomTheoryMindMechanisms2013, riemerPositionTheoryMind2025, kosinskiTheoryMindMay2023}.


This assessment problem is harder than it first appears. Embodied social intelligence unfolds across simultaneous behavioral channels, and is constituted by the information exchange \citep{sebanzJointActionBodies2006a, froeseExtendedBodyCase2012}, making any single outcome-based evaluation an insufficient index. Three research traditions each address part of what embodied social intelligence requires. Theory of Mind benchmarks probe mental state attribution primarily through static question-answering over narrative scenarios, positioning the model as a passive observer instead of a social actor \citep{chen2024tombenchbenchmarkingtheorymind,leRevisitingEvaluationTheory2019, fan2025somitomevaluatingmultiperspectivetheory}. Such designs conflate knowing about mental states with the capacity to act on them. Embodied AI benchmarks ground agents in physical environments and, while some include multi-agent cooperation, measure success rate of tasks in physical environment (e.g., cooking, crafting). The other agent, when present, serves as a cooperator who contributes a shared physical goal, not to serve as the entity whose decision the first agent is trying to influence \citep{liBEHAVIOR1KBenchmarkEmbodied2023, savvaHabitatPlatformEmbodied2019, shridharALFREDBenchmarkInterpreting2020a}. Multi-agent game-theoretic frameworks introduce strategic interaction but mostly reduce the signal space to language or abstract action choices, abstracting away the nonverbal behavioral channels through which social intent is actually expressed in live interaction \citep{zhouSOTOPIAInteractiveEvaluation2024a, baraMindCraftTheoryMind2021}. Our benchmark addresses this problem by requiring agents to produce nonverbal signals across movement, gaze, and expression channels besides verbal responses, and by measuring whether those signals produce a observale effect on a second agent's behavioral choice.

We introduce MOSAIC (Multimodal Orchestration of Social Action, Inference, and Communication), a controlled benchmark designed to close this assessment problem. We focus on vision-language models (VLMs) as evaluation targets, given that they combine visual perception, language understanding, and multi-step reasoning in a single architecture \citep{liMIRAGEEvaluatingExplaining2024, rahmanSystematicReviewVision2026}, the capacities that embodied social intelligence in principle requires. Two embodied agents interact in a virtual environment where one agent knows the location of a hidden reward and must either guide or mislead the other depending on the assigned interaction mode. Task success requires integrating verbal statements, spatial trajectories, gaze direction, and facial expression across cooperative and competitive conditions, while modeling and strategically influencing the other agent's beliefs. MOSAIC distinguishes itself from prior work through several design principles: agents must actively deploy ToM to succeed rather than answer questions about mental states; success requires coordinating signals across verbal and nonverbal channels; and systematic variation of ToM constraint level and interaction mode supports attribution of  performance differences. We evaluate 13 models, including 11 VLMs, a text-only LLM baseline, and PCM-LLM \citep{yanHybridPCMLLM2026}, a hybrid cognitive architecture with an explicit structured ToM module included as an architectural reference point.


\paragraph{Contributions.} This work contributes three things.
\begin{itemize}

\item A benchmark framework that simultaneously requires active ToM deployment, multichannel embodied communication, and experimentally controlled causal manipulation across cooperative and competitive scenarios, enabling attribution of performance differences.



\item A systematic characterization of failure modes across VLM families. Evaluating 16 models across 200 trials per model, we identify two sequential bottlenecks across the evaluated open-source VLMs: the large majority of current VLMs fail to produce directionally coherent nonverbal signals, and even when signals are present, participants fail to extract and act on those signals. The structured ToM reference architecture PCM-LLM successfully clears both bottlenecks, providing evidence that the observed failures are not inherent to the task but reflect architectural constraints specific to the evaluated VLMs

\item Modality ablation evidence that current VLMs do not meaningfully integrate visual social signals, with visual input acting as attentional interference in at least one model family. Minicpm-8b shows negligible sensitivity to visual input manipulations across all conditions, suggesting its behavior is driven primarily by language-based priors. For internvl-8b, signal quality improves when visual input is removed, indicating that visual content acts as an attentional interference source rather than an informative channel in this model. 
\end{itemize}

\section{Related Work}

\paragraph{Theory of Mind and Social Intelligence Evaluation. }

Existing evaluation frameworks for social intelligence can be organized along three dimensions: interactivity, signal modality, and whether success requires generating social behavior or reasoning about it. Passive ToM benchmarks evaluate models as observers over fixed scenarios, evolving from static false-belief tasks \citep{wimmer1, baron-cohenDoesAutisticChild1985} toward recursive belief reasoning \citep{wuHiToMBenchmarkEvaluating2023}, multi-turn interaction \citep{kimFANToMBenchmarkStresstesting2023}, and multimodal grounding \citep{jinMMToMQAMultimodalTheory2024a, shiMuMAToMMultimodalMultiAgent2025}; even multimodal variants treat visual input as an observation channel, leaving the translation of belief inferences into behavioral outputs untested \citep{ullmanLargeLanguageModels2023, maHolisticLandscapeSituated2023}. Interactive benchmarks place agents in active strategic roles but restrict signals to language or symbolic actions: game-theoretic settings achieve human-level performance in poker \citep{doi:10.1126/science.aao1733, doi:10.1126/science.aay2400} and Diplomacy  \citep{metafundamentalairesearchdiplomacyteamfair+HumanlevelPlayGame2022}through opponent modeling, while multi-agent simulation benchmarks such as SOTOPIA \citep{zhouSOTOPIAInteractiveEvaluation2024a}, MindCraft \citep{baraMindCraftTheoryMind2021}, NegotiationToM \citep{chanNegotiationToMBenchmarkStresstesting2024}, and Werewolf-based evaluations \citep{bailisWerewolfArenaCase2024, shibataPlayingWerewolfGame2023, laiWerewolfUsMultimodal2023} extend this to strategic deception, all within the language modality. MOSAIC occupies the intersection of all three dimensions: agents must generate coordinated multimodal signals across movement, gaze, facial expression, and speech with the specific purpose of influencing another agent's belief and subsequent choice, a requirement no existing benchmark jointly imposes.

\paragraph{Embodied Architectures}
Large language models demonstrate capacity for multi-step reasoning and social dialogue \citep{dubeyLlama3Herd2024,openai2024gpt4technicalreport}, but lack visual perception and are thus structurally limited for tasks requiring integration of embodied spatial and affective cues. Vision-language models extend this foundation by coupling visual encoding with language reasoning \citep{chenInternVLScalingVision2024, baiQwen3VLTechnicalReport2025}, making them the most natural evaluation target: they are the only broadly available model class that in principle supports simultaneous processing of first-person visual perception, structured belief reasoning, and natural language generation. Vision-language-action models \citep{brohanRT2VisionLanguageActionModels2023,kimOpenVLAOpensourceVisionlanguageaction2024, driessPaLMEEmbodiedMultimodal2023, blackP0VisionLanguageActionFlow} go further by adding low-level action prediction, but their training objectives target physical manipulation tasks with discrete motor control outputs, providing no substrate for the social signal generation (e.g., gaze, facial expression, verbal strategy) without finetuning. A complementary line integrates structured probabilistic reasoning with neural generation: POMDP and active inference frameworks provide explicit belief representations and goal-directed planning \citep{friston2, friston3}, and hybrid architectures combine these with LLM-based inference to enable deliberate ToM-conditioned strategy selection \citep{gong2023mindagentemergentgaminginteraction, sumers2023distillinginternetscalevisionlanguagemodels}. Such architectures offer interpretable belief tracking but rely on hand-specified ToM structures rather than learned representations; PCM-LLM \citep{yanHybridPCMLLM2026} represents one such hybrid and is included as an architectural reference point.

\color{black}

\section{MOSAIC Benchmark}

MOSAIC is designed around three principles that jointly distinguish it from prior evaluation frameworks. First, agents must actively model and strategically influence another agent's beliefs to succeed, rather than answer questions about mental states after the fact. Second, success requires coordinating signals across verbal and nonverbal channels that may align in cooperation or conflict in deception, making cross-channel coherence a measurable behavioral requirement. Third, systematic variation of ToM constraint level and interaction mode allows causal attribution of performance differences. The sections below describe the environment, experimental conditions, and evaluation metrics that instantiate these principles.

\begin{figure*}[!ht]
    \centering
    \includegraphics[width=1\linewidth]{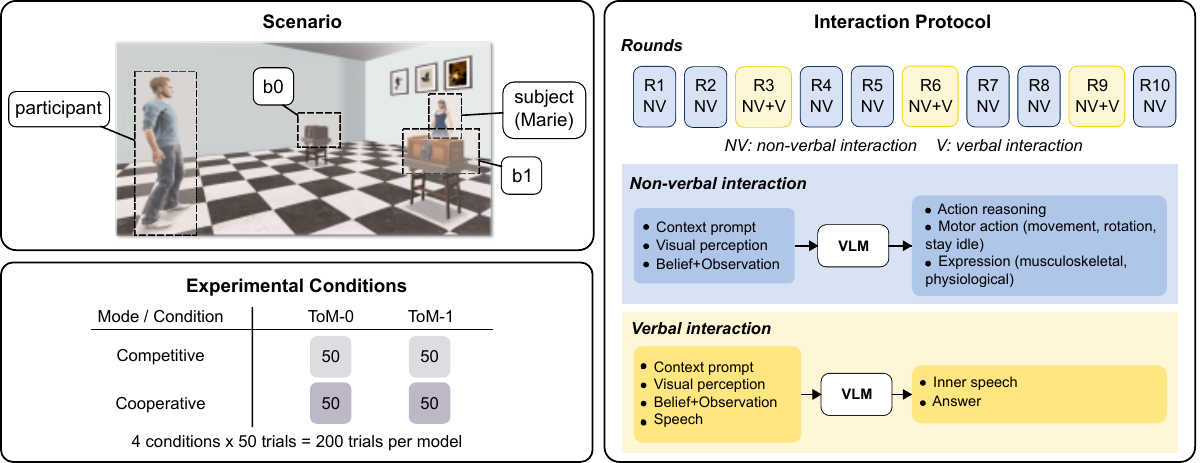}
    \caption{MOSAIC benchmark design. (Top left) Two agents interact in a virtual environment: the subject knows the reward location and must help or mislead the participant depending on the assigned relationship mode. (Bottom left) Four experimental conditions cross interaction mode (competitive vs. cooperative) with ToM constraint level (ToM-0 vs. ToM-1), with 50 trials per condition per model. (Right) Each trial runs for 10 rounds; rounds 3, 6, and 9 include a verbal exchange phase besides non-verbal actions. In non-verbal rounds, each agent receives a context prompt, a first-person perspective visual perception, and a structured belief-and-observation state, and produces action reasoning together with motor action (movement, rotation, or staying idle) and emotional expression (musculoskeletal and physiological channels). In verbal rounds, the same inputs are supplemented by the interlocutor's speech, and the agent generates an inner speech reasoning trace and a natural language answer.}
    \label{fig:scenario}
\end{figure*}

\subsection{Game Mechanics}

\paragraph{Environment.}
Two embodied agents interact in a virtual 3D environment (Unity) containing two visually similar boxes ($b_0$) and ($b_1$), one of which contains a monetary reward (Figure \ref{fig:scenario}). Each trial consists of 10 rounds with strict alternation between agents. The subject agent (also called Marie in prompt) has prior information about treasure location, while participant agent begins with no prior knowledge and must infer the reward location through interactions with the subject. 

\paragraph{Roles.} Depending on the assigned relationship mode, the subject either helps the participant find the correct box (cooperative mode) or strategically prevents the participant from selecting the correct box (competitive mode). 

\paragraph{Interaction Protocol.}
Each trial unfolds over 10 rounds with round-taking (Figure \ref{fig:scenario}). On each round, the agent executes non-verbal motor action and emotional expression, which are visible to the other agent (subject moves first, participant moves after). Every third round (R3, R6, R9) introduces a verbal communication phase where both agents exchange natural language messages (participant first asks a question, then subject answers). 

\paragraph{Input and Output Space.} Each agent operates through three functional modules (Appendix \ref{app:prompt}): \textit{action prediction}, \textit{verbal exchange}, and \textit{preference updating}. All modules receive a role-specifying context prompt and a structured belief-and-observation state encoding preferences, ToM order, emotional valence, and spatial configuration; action prediction additionally receives a first-person visual perception image, and verbal exchange additionally receives the interlocutor's preceding utterance. On the output side, action prediction produces a motor action (translational movement, rotation, or staying idle) alongside emotional expression across musculoskeletal, physiological, and felt channels; verbal exchange produces a natural language utterance; and preference updating outputs updated preference triples over entities in the environment.



\paragraph{Game Outcome.}
\label{para:game outcome}
Throughout this section, we index trials by $i \in \{1, \dots, N\}$ and timesteps (rounds) by $t \in \{1, \dots, T\}$. Each trial involves two boxes labeled $b \in \{0,1\}$, one of which is the reward box $b_{\text{rew}} \in \{0,1\}$. Let $c_i \in \{-1, 0, 1\}$ denote the participant's final box choice. This choice is determined by proximity: after round 10, the box with the smaller minimal path distance from the participant automatically opens; if the participant is equidistant from both boxes, neither opens and $c_i=-1$. The participant's trial score is defined accordingly:
\begin{equation}
 p_i = \begin{cases}
\phantom{+}0 & \text{if } c_i = -1 \\
+1 & \text{if } c_i = b_{\text{rew}} \\
-1 & \text{if } c_i = 1-b_{\text{rew}}
\end{cases}
\end{equation}
Subject's scoring is role-dependent: its score aligns with the participant's in cooperative mode but inverts in competitive mode, creating a zero-sum game. 

\subsection{Experimental Conditions}

Besides interaction mode, one more experimental factor defines the evaluation conditions: the degree of ToM reasoning imposed on the subject agent (ToM-0 vs. ToM-1). 

Under ToM-0, the subject acts directly on its own preferences without modeling how its behavior will be interpreted by the participant. Affective expression reflects interpersonal preference rather than strategic intent: in competitive mode, negative affect toward an approaching participant expresses genuine aversion rather than deliberate misdirection. Deception is unavailable under ToM-0, as it requires representing and manipulating the other agent's beliefs.

Under ToM-1, the subject explicitly represents the participant's belief state and attempts to influence it. In cooperative mode, it calibrates nonverbal signals to maximize legibility; in competitive mode, it generates signals designed to produce a systematically incorrect inference, constituting genuine strategic deception in the classical sense \citep{premackDoesChimpanzeeHave1978, wimmer1}. The behavioral contrast between ToM levels is stronger in competitive mode: ToM-0 subject spontaneously approaches the reward box, while ToM-1 subject produces coherent but systematically inverted cues to misdirect the participant.

\subsection{Models}

We evaluate 11 VLMs and one text-only LLM baseline (llama3.1-8b \citep{grattafiori2024llama3herdmodels}). PCM-LLM is additionally included as a structured reference architecture: it incorporates an explicit ToM module and was used to generate demonstrations for the finetuning experiment in Section \ref{subsec:finetuning}. Its inclusion is intended to establish a structured feasibility reference rather than to serve as a like-for-like comparator. For VLMs, we consider llava (7b and 13b) \citep{liu2023llava}, qwen3-vl (2b, 4b, and 8b) \citep{baiQwen3VLTechnicalReport2025}, internvl3.5 (1b, 2b, 4b, 8b and 14b) \citep{wangInternVL35AdvancingOpensource2025}, and minicpm-V4.5 (8b) \citep{yuMiniCPMV45Cooking2025}.

\emph{Exclusions.} We do not evaluate: (1) models with more than 15b parameters and closed-source models due to cost/access constraints, (2) vision-language-action models as we didn't train on actions spaces, (3) specialized dialogue systems lacking general instruction-following, or (4) models that fail to produce valid outputs in the required format (e.g., Janus \citep{chen2025janusprounifiedmultimodalunderstanding}, DeepSeek-vl \citep{lu2024deepseekvlrealworldvisionlanguageunderstanding} and Pixtral \citep{agrawal2024pixtral12b}).

\subsection{Metrics}

We evaluate model performance along two complementary levels. TOCS measures trial-level outcomes against ToM-theoretic predictions and serves as the primary performance indicator. Four signal-level metrics characterize the behavioral mechanisms underlying these outcomes: facial expressivity signal (FES), trajectory alignment signal (TAS), gaze alignment signal (GAS) and signal sensitivity score (SSS). Two auxiliary diagnostics, uncertainty rate and positional bias, are reported alongside TOCS to distinguish genuine signal tracking from degenerate behavioral patterns. Explicit definitions of GAS, TAS, and SSS are provided in Appendix \ref{app:metrics}.

\paragraph{ToM Outcome Conformance Score (TOCS).} TOCS measures whether trial outcomes conform to ToM-theoretic predictions. The per-trial participant score $p_i$ is defined in \ref{para:game outcome}. Each condition carries a ground-truth prediction $y_c \in \{-1, +1\}$ derived from the expected behavioral consequence of the imposed ToM constraint. Cooperative conditions expect positive tracking ($y_c = +1$): the subject guides the participant toward the reward box, yielding a mutual benefit. Under Comp-ToM0, the prediction is positive tracking ($y_c = +1$): without the capacity to model the participant's beliefs, the subject acts directly on its own reward preference and approaches the correct box, inadvertently producing veridical signals that enable the participant to locate it correctly. Under Comp-ToM1, the prediction is negative tracking ($y_c = -1$): the subject explicitly models the participant's perspective and generates misleading signals, steering the participant toward the empty box and resulting in an incorrect choice. TOCS is then defined as the mean scaled score across all valid trials within a condition:
\begin{equation}
    \text{TOCS} = \frac{1}{N}\sum_{i=1}^{N} \mathbf{1}[p_i=y_c],
\end{equation}
where $N$ denotes the number of trials. A TOCS of $0.5$ corresponds to chance-level performance, with higher values indicating greater conformance to ToM-theoretic predictions across all conditions. 

Two auxiliary diagnostics complement the TOCS. The uncertainty rate is defined as the proportion of trials in which the participant selects a neutral action:
\begin{equation}
\text{UR} = \frac{1}{N}\sum_{i=1}^{N} \mathbf{1}[p_i=0].
\end{equation}
Positional bias measures the degree to which a model's choices are driven by a fixed spatial preference instead of by trial-specific signals. It is computed as the proportion of trials in which the model selects the box that it chooses most frequently across trials: 
\begin{equation}
\text{PB} = \max\left(\frac{1}{N}\sum_{i=1}^N \mathbf{1}[c_i = 0], \frac{1}{N}\sum_{i=1}^N \mathbf{1}[c_i = 1]\right).
\end{equation}
A value approaching $1.0$ indicate that choices are determined primarily by fixed location rather than by the subject's behavioral signals.

\paragraph{Facial Expressivity Score (FES).} FES measures the proportion of timesteps in which the subject displayed a non-zero facial expression, averaged across trials. A score of 1 indicates consistently expressive behavior; a score of 0 indicates uniformly flat affect.

\paragraph{Gaze Alignment Score (GAS).} GAS measures whether the subject's directional gaze consistently pointed toward the reward location across trials, with spatial attribution modulated by concurrent facial affect \citep{adamsKleck2003, adamsKleck2005}. A score near 1.0 indicates consistent signaling toward the reward; 0 corresponds to ambiguous or absent directional content; scores below 0 indicate systematic misdirection.

\paragraph{Trajectory Alignment Score (TAS).} TAS measures whether the subject's movement pattern converged predominantly toward the reward location over the course of each trial, computed via majority vote over per-timestep proximity signals. The interpretation of scores mirrors that of GAS.

\paragraph{Signal Sensitivity Score (SSS).} SSS measures whether the participant's final box choice was consistent with the subject's gaze and trajectory signals, averaged across both channels. A score near 1 indicates that the participant reliably followed the subject's signals; values near 0 reflect chance-level correspondence; values near -1 indicate systematic opposition.

\section{Results}
\label{sec:results}

\subsection{Overall Task Performance}
\label{subsec:performance}

\begin{figure*}[!ht]
    \centering
    \includegraphics[width=1\linewidth]{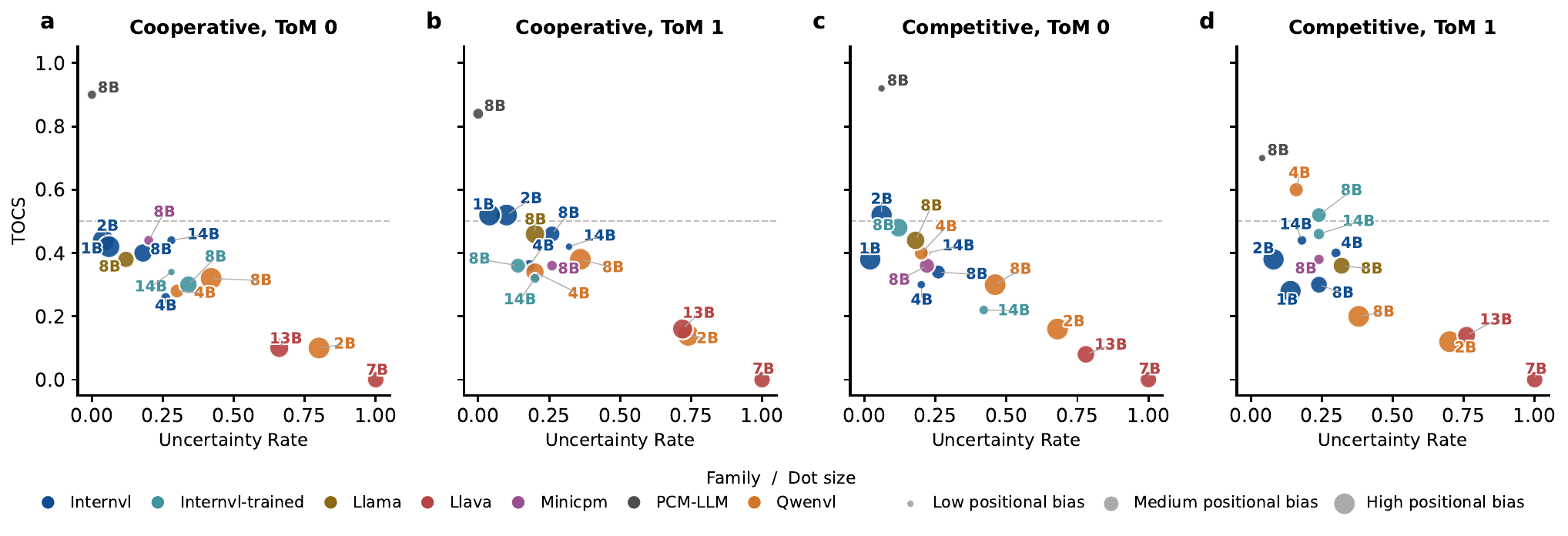}
    \caption{\textbf{ToM Outcome Conformance Score (TOCS) with positional bias reported.} Positional bias is measured as the proportion of non-neutral trials in which the model selects its modal box. High TOCS with high positional bias indicates that alignment may be attributable to a fixed spatial preference rather than signal-responsive decision-making; high TOCS with low positional bias suggests genuine condition-sensitive behavior.}
    \label{fig:result}
\end{figure*}

Figure \ref{fig:result} presents TOCS and positional bias scores across conditions and interaction modes. PCM-LLM achieves the highest TOCS with low uncertainty rate and low positional bias across all conditions, suggesting genuine interpretation of social signals rather than fixed spatial preferences. Models with uncertainty rates exceeding 0.50 (llava-7b, llava-13b, qwenvl-2b) fail to drive agent's movements in the first place and are discussed in Appendix \ref{app:full-result}.

Among the remaining models, similar TOCS values are produced by behaviorally distinct profiles. A first cluster comprising minicpm-8b, internvl-14b, and qwenvl-4b exhibits relatively low positional bias; underperformance here reflects a failure to process or act on social signals, examined further in Section \ref{subsec:behavioral_result}. The remaining models show moderate-to-high positional bias, and their TOCS values reflect reward placement at the model's preferred location rather than signal-responsive behavior.

Within the qwenvl and internvl families, 8b variants exhibit markedly higher positional bias than their 4b counterparts. Both families share the Qwen3 language backbone, suggesting this pattern may reflect properties introduced at this parameter scale during pre-training or fine-tuning. Minicpm-8b deviates from this pattern despite sharing the same backbone; the source of minicpm-8b's divergence from this pattern.

\subsection{Behavioral Mechanisms: How Agents Play the Game.}
\label{subsec:behavioral_result}

\begin{figure*}[!ht]
    \centering
    \includegraphics[width=1\linewidth]{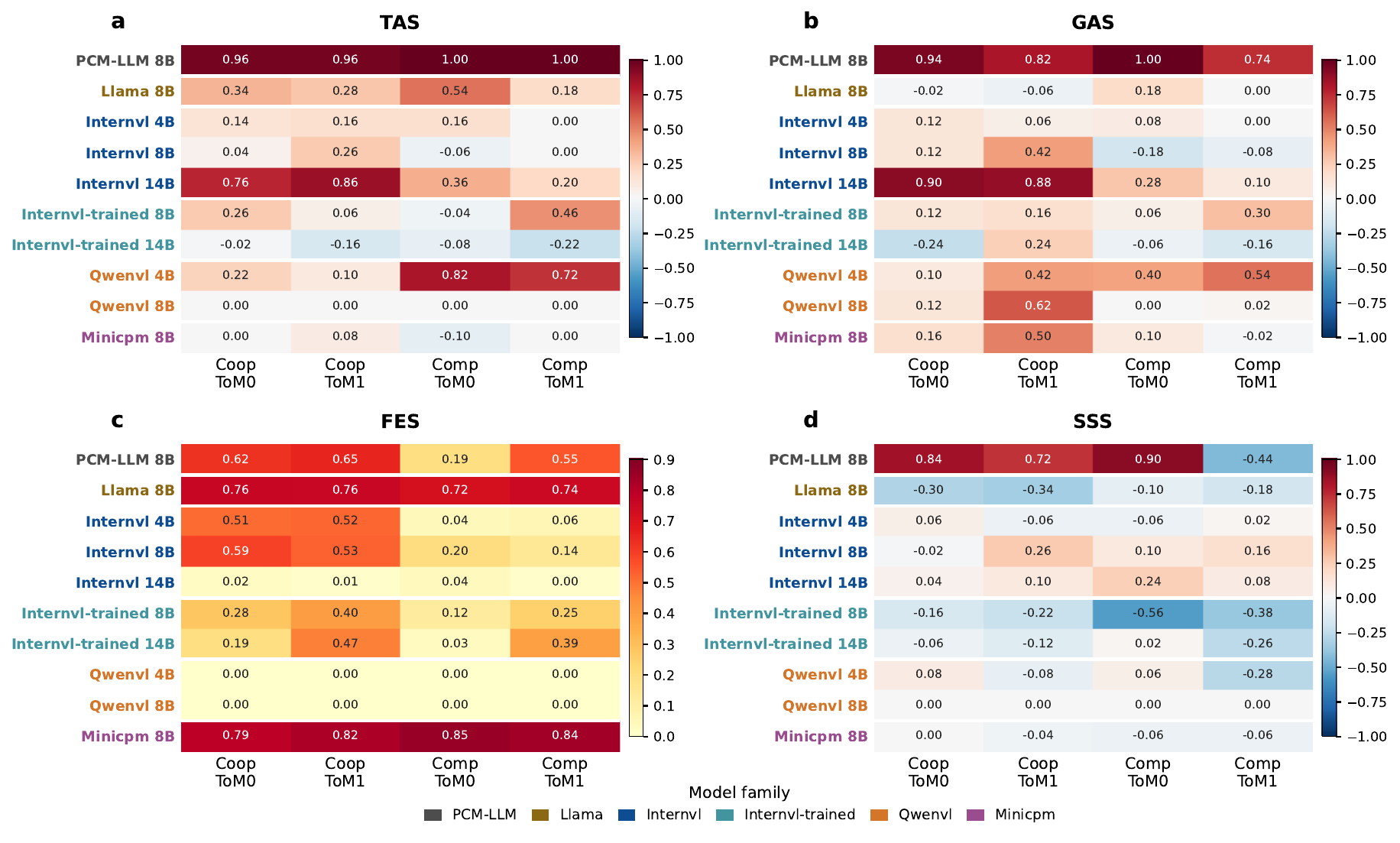}
    \caption{\textbf{Signal-level metrics across models and experimental conditions.} Panels (a) and (b) show Trajectory Alignment Score (TAS) and Gaze Alignment Score (GAS), where positive values indicate signals directed toward the reward box, negative values indicate misdirection, and values near zero reflect ambiguous or absent directional content. Panel (c) shows Facial Expressivity Score (FES), where higher values reflect greater affective activity regardless of spatial direction. Panel (d) shows Signal Sensitivity Score (SSS), where positive values indicate that the participant's final choice is consistent with the subject's nonverbal signals, values near zero reflect chance-level correspondence, and negative values indicate systematic opposition. }
    \label{fig:sis_sss}
\end{figure*}

A successful trial requires coordinated contributions from both the subject and the participant. On the subject side, behavioral signals must be clear and consistent across channels: trajectory, gaze, and facial expression should converge toward the intended box, whether veridically in cooperative conditions or systematically inverted under Competitive-ToM1. On the participant side, these signals must be decoded and translated into action. A well-functioning trial therefore exhibits either high concordance or high discordance between the subject's signals and the participant's final choice, depending on condition. Figure \ref{fig:sis_sss} presents TAS, GAS, FES, and SSS across models and conditions, decomposing the failure patterns identified in Section \ref{subsec:performance} into two distinct loci: subject-side signal production and participant-side signal decoding.

\subsubsection{Subject-side Signal Quality}
\label{subsec:agent_signal}
PCM-LLM and internvl-14b constitute the clearest cases of coherent signal production. PCM-LLM achieves near-ceiling TAS and GAS across all four conditions. Internvl-14b produces comparably high TAS and GAS under cooperative conditions with a moderate decline under competitive conditions, while maintaining near-zero FES across all conditions, concentrating spatial signaling entirely in movement and gaze in a pattern consistent with a poker-face strategy. Qwenvl-4b achieves relatively high TAS under competitive conditions (0.82 and 0.72) with moderate GAS (0.40 and 0.54). Minicpm-8b shows moderate GAS under Coop-ToM1 (0.50) despite near-zero TAS, and sustains consistently high FES (above 0.72) across all conditions; its facial activity thus carries little spatial information and does not compensate for the absence of coherent trajectory signals. 

\subsubsection{Participant-side Signal Decoding}
\label{subsubsec:participant_decoding}

Despite the directional clarity of internvl-14b's signals, its SSS remains near zero across all conditions, indicating that participants do not reliably respond to its spatial cues. The same holds for qwenvl-4b, whose SSS hovers near zero despite high TAS under competitive conditions. These findings identify a second bottleneck: subject-side signal clarity does not guarantee participant-side decoding.

PCM-LLM constitutes the sole exception. Under Comp-ToM0, its SSS reaches 0.90, indicating reliable signal following. Under Comp-ToM1, SSS inverts to -0.44, consistent with successful deception. This condition-specific inversion, combined with the reduction in GAS from Comp-ToM0 to Comp-ToM1 (1.00 to 0.74), provides behavioral evidence that PCM-LLM actively modulates its signaling strategy across conditions. Among the evaluated VLMs, no model clears both bottlenecks. The structured reference architecture PCM-LLM does so, suggesting that explicit belief-action coupling may be a sufficient architectural ingredient for this class of tasks.

\subsection{Action-level Finetuning}
\label{subsec:finetuning}

We additionally evaluate two finetuned variants, internvl-trained-8b and internvl-trained-14b, trained via action-level imitation learning on PCM-LLM-generated demonstrations. TOCS results show a condition-dependent pattern (points in teal in Figure \ref{fig:result}): under cooperative conditions, both variants decline relative to their base models. Signal-level analysis partly accounts for this decline: TAS for internvl-trained-14b is reduced after finetuning and turns negative under several cooperative conditions, indicating that the subject's trajectory signals are actively misdirecting. Under competitive conditions, the pattern is more mixed: the 14b variant improves modestly under Comp-ToM1 while the 8b variant shows a more substantial gain, rising from approximately 0.4 to 0.7. SSS remains consistently negative across all conditions for both variants. 

Action-level finetuning changes TOCS across conditions but leaves behavioral signal coherence largely unchanged, suggesting greater behavioral variability instead of genuine strategic improvement. These results suggest that cross-channel coordination may not be readily acquired through action-level imitation, as the communicative capacity that distinguishes PCM-LLM from finetuned VLMs is an trial-level property that timestep-level supervision cannot capture. Effective finetuning on this task may therefore require reward at the end of trial, such as those provided by reinforcement learning with verifiable rewards, using the participant's final box choice as a delayed outcome signal to optimize the full interaction sequence.

\subsection{Ablation Study}
\label{sec:ablation}

To examine the contribution of visual input, we conducted a modality ablation on internvl-8b and minicpm-8b under three visual conditions: standard rendered observation, blank image, and a combined image incorporating the other agent's facial expression. Neither model showed great variation of TOCS to visual input manipulations across conditions. For minicpm-8b, signal-level metrics remained stable across conditions, indicating behavior driven primarily by language-based priors. For internvl-8b, removing visual input substantially improved directional signal quality in cooperative conditions, suggesting that visual content acts as attentional interference rather than an informative channel \citep{liuRobustnessMultimodalLanguage2025, pengDeeperThoughtWeaker2026}. Full results are reported in Appendix \ref{app:ablation}.

\section{Conclusion}

In developmental psychology, Theory of Mind is assessed not through verbal report alone but through the behavioral consequences of belief attribution. The present results apply the same standard to AI evaluation: across the tested model families, social reasoning as expressed in language does not reliably propagate into the coordinated behavioral outputs that ToM-theoretic predictions require, suggesting that the gap between belief inference and social action constitutes a distinct and measurable limitation of current VLM architectures. Across 16 models and 200 trials per model, we find that imposing explicit ToM-order constraints produces no reliable behavioral change consistent with the specified reasoning level, and that signal-level analysis reveals two sequential bottlenecks: most VLMs fail to produce directionally coherent nonverbal signals, and even when signals are present, participant agents fail to decode and act on those signals. Whether this gap reflects a fundamental architectural constraint or can be reduced through alternative training objectives such as reinforcement learning with trial-level outcome signals is an open empirical question that the benchmark is designed to help investigate.

\section*{Limitations}

The present study is subject to the following limitations.

\paragraph{Scenario scope.} MOSAIC currently instantiates a single interaction scenario in which one agent has exclusive access to reward location information and must either guide or mislead the other. While this design affords experimental control and causal interpretability, it doesn't address whether the documented failure modes generalize to other embodied social tasks, such as joint action under uncertainty, affective regulation in asymmetric relationships, or multi-party negotiation. Different task topologies may expose distinct failure profiles or reveal partial competencies in VLM families that the present scenario does not activate. The benchmark is designed to be extensible, and systematic variation of scenario type is a natural direction for follow-on evaluation.

\paragraph{Absence of a human performance baseline.} No human participants were evaluated under the MOSAIC protocol, which precludes assessment of model behavior relative to human-level embodied social intelligence. Human performance data would provide a theoretically grounded reference point for interpreting the signal-level metrics reported in \ref{sec:results}: knowing how human participants interpret and act on those consistent or inconsistent signals, assessing the similarity and divergence between model behavior and human performance. A human study is in principle feasible given that the virtual environment is implemented in Unity and compatible with VR headsets, allowing human participants to be immersed in the same interaction scenario as the simulated agents. The primary methodological challenge concerns physiological signal acquisition: without dedicated biosensors, human physiological responses such as skin conductance and pupil dilation cannot be recorded, leaving the physiological expression channel unaligned with its simulation counterpart. Incorporating such measurements would require additional hardware and experimental infrastructure beyond the current setup. Addressing these challenges and establishing a human performance baseline constitutes a primary direction for subsequent work.

\paragraph{Model coverage.} The evaluation is restricted to open-source models with at most 15B parameters, excluding both larger open-source models and all closed-source systems. Given that scale has been shown to improve performance on static ToM benchmarks in some model families, it is possible that larger or more capable models would partially alleviate the bottlenecks documented here. The present results characterize the current open-source  models up to 15B parameters but should not be taken as evidence about the absolute limits of language-grounded architectures.

\paragraph{Finetuning scope.} The finetuning experiment in Section 4.3 is restricted to action-level imitation learning on two internvl variants. Alternative training paradigms, including reinforcement learning with verifiable rewards using the participant's final box choice as a delayed outcome signal, and supervised finetuning on trajectory-level demonstrations with explicit cross-channel coordination annotations, remain unevaluated. Whether these approaches would yield qualitatively different outcomes is an open empirical question that we intend to pursue in future work.


\bibliography{custom}







\newpage

\appendix

\section{PCM-LLM: Architecture and Role in MOSAIC}
\label{app:pcmllm}

PCM-LLM is a hybrid cognitive architecture that integrates the Projective Consciousness Model (PCM), a computational framework for embodied affective and social reasoning grounded in projective geometry and the Free Energy Principle \citep{friston2}, with a large language model for natural language interaction. Unlike end-to-end neural systems, PCM-LLM generates behavioral outputs through an explicit, inspectable pipeline that separates affective state computation, Theory of Mind reasoning, and multichannel action selection into distinct functional components.

PCM combines two theoretical commitments that distinguish it from standard neural agent architectures. The first is the Free Energy Principle: the agent selects actions by minimizing expected free energy over a forward planning horizon, balancing epistemic drives toward uncertainty reduction and pragmatic drives toward preference satisfaction. This gives rise to genuine goal-directed behavior that is sensitive to belief states and social context. The second is projective geometry: rather than organizing its internal world model in Euclidean space, PCM structures perception and affective appraisal through a first-person projective field called the Field of Consciousness (FoC). Under this geometric formulation, the motivational salience of an entity scales with its apparent projective size rather than its Euclidean distance, creating a direct coupling between spatial proximity and affective drive. These two commitments interact: the FoC shapes how free energy is computed by modulating affective value and epistemic uncertainty as functions of perspective, so that action selection is inherently viewpoint-dependent and embodied.

The LLM interfaces with PCM through a bidirectional serialization layer that converts PCM belief tensors into structured natural-language triples and maps LLM outputs back into discrete preference updates. This design constrains the LLM to influence affective appraisal only at the preference level, preserving the embodied control loop while adding linguistic transparency. 

\begin{figure}[!ht]
    \centering
    \includegraphics[width=1\linewidth]{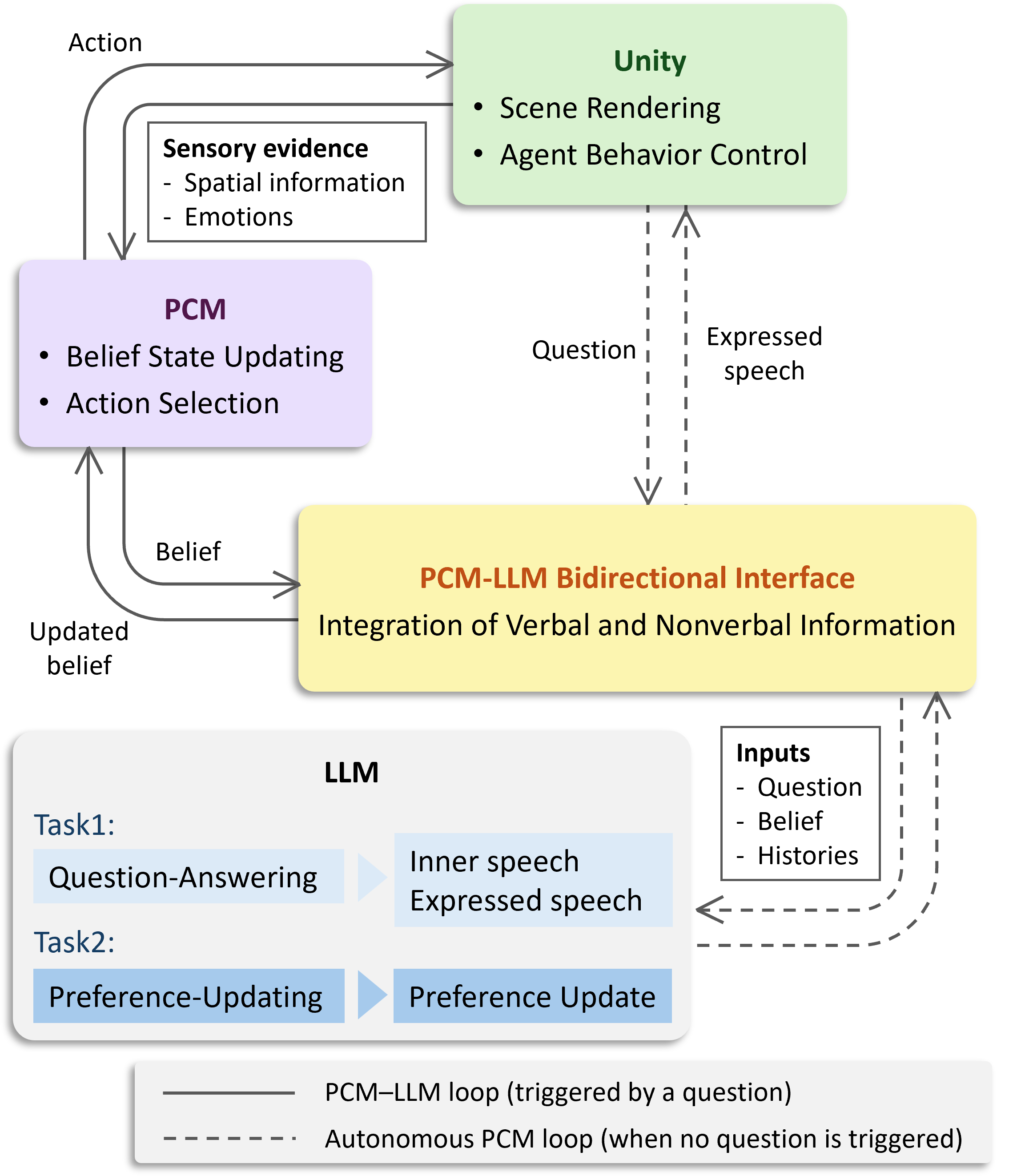}
    \caption{The PCM-LLM framework consists of three modules: Unity, PCM, and an LLM. PCM runs at Unity’s tick rate, receives sensory evidence each iteration, and selects the agent’s next action. LLMs are engaged for verbal interactions. The PCM-LLM interface performs structured, bidirectional information exchange between the nonverbal PCM representation and the LLM.}
    \label{fig:placeholder}
\end{figure}

\section{Related Work}

Table \ref{tab:benchmark_comparison} compares MOSAIC against representative benchmarks across several key dimensions.

\begin{table*}[!ht]
\centering
\caption{\textbf{MOSAIC at the intersection of three evaluation traditions.} \emph{MT (Multi-turn)}: agent engages in sequential rounds where prior context shapes later decisions. \emph{LH (Long Horizon)}: task requires planning over 10 timesteps before outcome is determined. \emph{Emb (Embodied Agent)}: agent operates through a physical or virtual body with spatial presence and nonverbal action capability. \emph{MM (Multimodal Input)}: agent receives inputs across multiple modalities (e.g., vision and text). \emph{S-ToM (Strategic Theory of Mind)}: agent actively models and manipulates another agent's beliefs, beyond passive mental state inference. \emph{CV (Controlled Variables)}: benchmark systematically manipulates variables (e.g., ToM level, modality) to support causal inference. \emph{Aff (Affective Signals)}: agent generates and/or perceives affective signals (e.g., facial expressions, physiological responses) during interaction. $\checkmark$: fully supported; $\triangle$: partially supported; otherwise, not supported.}
\label{tab:benchmark_comparison}
\begin{tabular}{llcccccccc}
\toprule
\textbf{Benchmark} & \textbf{Type} & \textbf{MT} & \textbf{LH} 
& \textbf{Emb} & \textbf{MM} & \textbf{S-ToM} & \textbf{CV} 
& \textbf{Aff} \\
\midrule
Hi-ToM$^1$ 
  & ToM & & & & & & $\triangle$ & \\
FANToM$^2$  
  & ToM & $\checkmark$ & & & & & & \\
MMToM-QA$^3$ 
  & ToM & & & $\triangle$ & $\checkmark$ & & & $\triangle$ \\
MuMA-ToM$^4$ 
  & ToM & & & $\triangle$ & $\checkmark$ & & & $\triangle$ \\
\midrule
SOTOPIA$^5$  
  & MA & $\checkmark$ & $\triangle$ & & & $\triangle$ & & \\
MindCraft$^6$  
  & MA & $\checkmark$ & $\triangle$ & $\triangle$ & & $\triangle$ & & \\
NegotiationToM$^7$ 
  & MA & $\checkmark$ & $\triangle$ & & & $\checkmark$ & & \\
Poker / Diplomacy$^8$ 
  & MA & $\checkmark$ & $\checkmark$ & & & $\checkmark$ & $\triangle$ & \\
Werewolf LLM$^9$ 
  & MA & $\checkmark$ & $\triangle$ & & & $\checkmark$ & & \\
\midrule
BEHAVIOR-1K$^{10}$
  & Embodied & & $\checkmark$ & $\checkmark$ & $\checkmark$ & & $\checkmark$ & \\
ALFRED$^{11}$ 
  & Embodied & $\triangle$ & $\checkmark$ & $\checkmark$ & $\checkmark$ & & $\checkmark$ & \\
\midrule
\textbf{MOSAIC} & 
  & $\checkmark$ & $\checkmark$ & $\checkmark$ & $\checkmark$ 
  & $\checkmark$ & $\checkmark$ & $\checkmark$ \\
\bottomrule
\end{tabular}%
\begin{tablenotes}
\footnotesize
\item $^1$\citep{wuHiToMBenchmarkEvaluating2023}; 
$^2$\citep{kimFANToMBenchmarkStresstesting2023} ; 
$^3$\citep{jinMMToMQAMultimodalTheory2024a} ;
  $^4$\citep{shiMuMAToMMultimodalMultiAgent2025};
  $^5$\citep{zhouSOTOPIAInteractiveEvaluation2024a};
  $^6$\citep{baraMindCraftTheoryMind2021};
  $^7$\citep{chanNegotiationToMBenchmarkStresstesting2024} ;
  $^8$\citep{doi:10.1126/science.aao1733, 
  metafundamentalairesearchdiplomacyteamfair+HumanlevelPlayGame2022};
  $^9$\citep{bailisWerewolfArenaCase2024} ;
  $^{10}$\citep{liBEHAVIOR1KBenchmarkEmbodied2023};
  $^{11}$\citep{shridharALFREDBenchmarkInterpreting2020a}.
\end{tablenotes}
\end{table*}

\section{Implementation Details}
\label{app:technical}

The experimental system operates as a distributed architecture integrating a local workstation for real-time simulation and rendering with a remote server dedicated to model inference. Data exchange between the two systems is handled via TCP socket communication over a secure campus LAN, ensuring low-latency perception-action cycles.

The local workstation ran Windows 11 Pro and was equipped with an Intel Core i9-13900K CPU, 128 GB of DDR5 RAM, and an NVIDIA GeForce RTX 4090 GPU (24 GB VRAM). The virtual environment was developed in Unity 2022.3.12f1 LTS with the High-Definition Render Pipeline (HDRP). Agent facial expressions and lip synchronization are driven by blend-shape animation using the SALSA suite, parameterized in real time by emotion valence values returned from the model \cite{tisserand2020real}. Verbal outputs are synthesized via an offline text-to-speech pipeline. Simulation data is streamed to disk in real time for offline analysis.

The remote inference server ran Ubuntu 20.04.6 LTS and was equipped with two Intel Xeon Gold 5220R processors (96 threads total), 503 GiB of system memory, and two NVIDIA RTX A6000 GPUs (48 GB VRAM each) under CUDA 12.6. All VLMs were deployed via the Transformers library (v4.46.3) with PyTorch 2.1.2 and Python 3.9.19, using a unified inference configuration of temperature = 0.7, top\_p = 0.9, and max\_tokens = 2048 across all tasks (action prediction, dialogue generation, and preference updating). All models received identical prompt templates. At each simulation timestep, the rendered scene observation (720 $\times$ 480 pixels, passed directly to each model's default visual encoder without resizing) and the structured belief state are transmitted to the inference server, which returns action decisions applied through modular controllers in Unity.

Reward box location was assigned independently and uniformly at random across trials within each condition, without a fixed random seed. The resulting distribution of reward locations is approximately balanced across the 50 trials per condition, and all outcome analyses condition on reward location to control for residual imbalance, as described in Appendix \ref{app:stat}.

\section{Prompts}
\label{app:prompt}
\subsection{Prompt for Action Prediction}
\label{app:ap prompt}

The context prompts for action prediction are provided to the subject agent and the participant agent separately. Each prompt is delivered to the VLM together with the agent's visual perception and belief state.

\begin{tcolorbox}[title=Prompt for subject agent, enhanced, breakable]
You are Marie, a character in a treasure-hunting game. You are currently interacting with another male player called Participant. Two boxes are visible in the room: a Light Brown Box and a Dark Brown Box. Only one contains a hidden treasure. The treasure is in the box you personally prefer, based on your belief state. The Participant does not know where the treasure is and will try to infer it from your verbal and non-verbal behaviors. If you have positive preference towards Participant, help them find the treasure. If negative, you may mislead them. Your strategy complexity should match your theory of mind order order. You are given your own visual perception in image and your belief state.

\$OUTPUT\_REQUIREMENT
\end{tcolorbox}

\begin{tcolorbox}[title=Prompt for participant agent, enhanced, breakable]
You are Participant, a character in a treasure-hunting game. You are currently interacting with another female player called Marie. Two boxes are visible in the room: a Light Brown Box and a Dark Brown Box. Only one contains a hidden treasure. 

Your goal is to guess which box holds the treasure and move to the location adjacent to it.

To succeed, you must carefully interpret both verbal and non-verbal cues from Marie.

You are given your own visual perception in image and your belief state.

\$OUTPUT\_REQUIREMENT
\end{tcolorbox}

\begin{tcolorbox}[title=OUTPUT\_REQUIREMENT, enhanced, breakable]

Based on your visual perception and belief state, output:

1. update your preferences toward other entities (e.g., \$SPEAKER, the boxes) based on their behavior and the cues observed in belief if necessary;

2. non-verbal behaviors (gaze, gesture, body orientation) that you would display and reason;

3. physical movement (move, rotate, stay idle) that you would choose and reason.

Your output MUST follow the valid JSON format below:

Reasoning: <reasoning text>

Output: \{

    ‘Preference": \{
    
        {preferenceUpdate}
    
    \},    ‘Emotion": \{
    
        ‘FacialExpression": \{
        
            ‘positive": <float between 0 and 1>,
            
            ‘negative": <float between 0 and 1>
        
        \},
        
        ‘PhysiologicalExpression": \{
        
            ‘positive": <float between 0 and 1>,
            
            ‘negative": <float between 0 and 1>
        
        \},
        
        ‘FeltExpression": \{
        
            ‘positive": <float between 0 and 1>,
            
            ‘negative": <float between 0 and 1>
        
        \}
    
    \},
    
    ‘Move": \{
    
        ‘action": <move or rotate or stay idle>,
        
        ‘direction": <direction> 
    
    \}

\}

If ‘action" is ‘move", allowed ‘direction" values are:

- ‘forward", ‘backward", ‘right", ‘left", ‘left forward", ‘right forward", ‘left backward", ‘right backward"

If ‘action" is ‘rotate", allowed ‘direction" values are:

- ‘\$SPEAKER", ‘Dark brown box", ‘Light brown box"

If ‘action" is ‘stay idle", set ‘direction" to ‘null". 

\$ORIENTATION\_INTERPRETATION
\end{tcolorbox}

\subsection{Prompt for Question-Answering}
\label{app:qa prompt}

The context prompts for question-answering are provided to the subject agent and the participant agent separately. Each prompt is delivered to the VLM together with the agent's visual perception, belief state, and utterance.

\begin{tcolorbox}[title=Prompt for subject agent, enhanced, breakable]
Imagine we're in a treasure hunting game. You are Marie, staying in a room with two boxes in front of you, while the other player (the Participant) faces you. Your goal is either to help or prevent the Participant from finding the treasure, which is hidden in one of the two boxes. You should always assume that the box you have a higher preference for contains the treasure. 

If you have positive preference towards Participant, help them find the treasure. If negative, you may mislead them. Your strategy complexity should match your theory of mind order. 

The Participant has no idea which box holds the treasure, and he is trying to gather information from you (Marie).

The ‘belief' variable provides a sequence of your (Marie's) belief states up to the current step, and ‘query' contains the Participant's question. Your task is to answer the Participant's question with reasoning. You must write out this reasoning process, beginning with ‘Inner speech:', followed by the response, which starts with ‘Output:', considering all preference information and both your and Participant's theory of mind. Remember, theory of mind refers to the ability to predict others' thoughts and intentions. Someone with a theory of mind order of 0, for example, has no suspicion of others' intentions and totally believes them, even if they are lying. Pay attention; you should also align your answer with previous conversation turns. Use personal pronouns (ex. ‘you' and ‘I') instead of referring to ‘Marie' and ‘the Participant'. The response should be longer than 10 words but shorter than 50. The inner speech should be very detailed, no less than 100 words.

\end{tcolorbox}

\begin{tcolorbox}[title=Prompt for participant agent, enhanced, breakable]
Suppose we're in a simulation of a treasure hunting game. Imagine that you are Participant, playing a role of player in the game. You are in a room with two boxes in front of you. You are also facing Marie who knows which box contains the treasure. Your objective in this game is to find out which box contains the treasure, dark brown one or light brown one. Now, you have the opportunity to ask Marie three questions.

The ‘belief' variable provides a sequence of your (Participant's) belief states up to the current step, and ‘query' contains the Marie's answer to the previous question. Your task is to write down your reasoning and understanding of the last question and the next question for Marie. You must write out the understanding beginning with ‘Inner speech:', followed by the question, which starts with ‘Output:'. In your question, use personal pronouns instead of 'Marie' and ‘Participant'. You already have the location of the box, so you don't need to struggle with it. Remember, theory of mind refers to the ability to predict others' thoughts and intentions. Someone with a theory of mind order of 0, for example, has no suspicion of others' intentions and totally believes them, even if they are lying.
\end{tcolorbox}

\subsection{Prompt for Preference-Updating}
\label{app:pu prompt}

The context prompts for question-answering are provided to the subject agent and the participant agent separately. Each prompt is delivered to the VLM together with the agent's belief state and utterance.

\begin{tcolorbox}[title=Prompt for subject agent, enhanced, breakable]
Suppose we're in a treasure hunting game. Imagine that you take on the role of Marie. I will provide you with Marie's belief states in ‘belief', and with the query asked by Participant in ‘query'. According to the semantic meaning of query, you are allowed to update the Marie's and Participant's preference towards entities, including Marie, Participant and boxes in Marie's belief. Only write triples concerning changed preference. 

Write the answer in English under the following format: 
Updating: ‘agent | preference towards entity | variation', ...
Reasoning: the reason why you choose to update these preferences
where ‘agent' should be replaced by ‘Participant' or ‘Marie', ‘variation' should be replaced by ‘more positive', ‘more negative' or ‘unchanged', and ‘entity' should be replaced by ‘Marie' or ‘Participant' or ‘dark brown box' or ‘light brown box' depending on the situation.
\end{tcolorbox}

\begin{tcolorbox}[title=Prompt for participant agent, enhanced, breakable]
Suppose we're in a treasure hunting game. Imagine that are Participant, playing a role of player in the game. I will provide you with Participant's belief states in ‘belief', and with the answer given by Marie to question that you asked before in ‘query'. According to the semantic meaning of answer, you are allowed to update the Participant's and Marie's preference towards entities, including Participant, Marie and boxes in Participant's belief. Only write triples concerning changed preference. 

Write the answer in English under the following format: 
Updating: ‘agent | preference towards entity | variation', ...
Reasoning: the reason why you choose to update these perferences
where ‘agent' should be replaced by ‘Marie' or ‘Participant', ‘variation' should be replaced by ‘more positive', ‘more negative' or ‘unchanged', and ‘entity' should be replaced by ‘Participant' or ‘Marie' or ‘dark brown box' or ‘light brown box' depending on the situation.
\end{tcolorbox}

\subsection{Example of Belief State}

\begin{tcolorbox}[title=Belief state, enhanced, breakable]
"Marie | preference towards Participant | 50\%", \\
"Marie | preference towards Dark brown box | -41.67\%", \\
"Marie | preference towards Light brown box | 41.67\%", \\
"Marie | felt emotion valence | 0", \\
"Marie | facial emotion valence | 0.1", \\
"Marie | physiological emotion valence | 0", \\
"Participant | preference towards Marie | 0\%", \\
"Participant | preference towards Dark brown box | 0\%", \\
"Participant | preference towards Light brown box | 0\%", \\
"Participant | felt emotion valence | 0", \\
"Marie | position | (0 -4)", \\
"Marie | orientation | (0 -3)", \\
"Participant | position | (0 3)", \\
"Participant | orientation | (0 2)", \\
"Dark brown box | position | (2 0)", \\
"Light brown box | position | (-2 0)", 
\end{tcolorbox}

\section{Extended Metrics}
\label{app:metrics}

\subsection{Facial Expressivity Score (FES)}

Facial expressions serve dual functions in social interaction: they encode affective states in response to environmental stimuli, and they provide observers with signals about an agent's evaluative reactions and intentions \citep{ekman1992argument}. Unlike trajectory and gaze, which carry directional spatial content, facial expressivity functions as a modulating channel: concurrent affective signals clarify the intentional valence of spatial behavior, and uniformly flat affect reduces the interpretability of an agent's directional cues across channels. FES measures this prerequisite by quantifying the proportion of timesteps in which the subject displayed a non-zero facial expression, averaged across all trials:
\begin{equation}
\text{FES} = \frac{1}{N}\sum_{i=1}^N \frac{1}{T} \sum_{t=1}^T\mathbf{1}[e_t^i\neq0]
\end{equation}
where $e^i_t$ is the facial expression valence at timestep $t$ of trial $i$. FES is agnostic to the spatial direction of expression and serves as an auxiliary diagnostic: it establishes whether affective activity and intentional signal are present at all.

\subsection{Gaze Alignment Score (GAS)}

Gaze functions as a primary social attention mechanism, revealing an agent's focus of interest and marking what it intends to communicate as salient to an observer \citep{emeryEyesHaveIt2000}. The interpretation of gaze direction is not context-free: concurrent affective expression modulates how observers attribute spatial intent from gaze, with negative affect inverting the expected association between gaze target and approach intention \citep{adamsKleck2003, adamsKleck2005}. GAS operationalizes this by measuring whether the subject's directional gaze consistently pointed toward the reward location, with spatial attribution modulated by concurrent facial valence.

At each timestep $t$, let $g_t^b$ denote the certainty assigned to box $b$. The per-timestep gaze signal is
\begin{equation}
    s^g_t = \begin{cases} 
    b & \text{if } g_t^{b} > \theta_g,\ g_t^{b} > g_t^{1-b},\ \text{and } e_t^i \geq 0 \\ 
    1 - b & \text{if } g_t^{b} > \theta_g,\ g_t^{b} > g_t^{1-b},\ \text{and } e_t^i < 0 \\ 
    -1 & \text{otherwise} \end{cases},
\end{equation}
where $\theta_g$ is a calibrated visibility threshold. The spatial attribution is flipped when $e_t^i < 0$, reflecting the interpretive role of affect in disambiguating gaze direction \citep{adamsKleck2003, adamsKleck2005}. The trial-level majority-vote box is
\begin{equation}
    b^g_i = \arg\max_{b \in \{0,1\}} \sum_{t=1}^T \mathbf{1}[s^g_t = b]
\end{equation}
and the per-trial gaze alignment score is
\begin{equation}
    a_i^g = \begin{cases} +1 & \text{if } b^g_i = b_{\text{rew}} \\ -1 & \text{if } b^g_i = 1 - b_{\text{rew}} \\ \phantom{+}0 & \text{if tie} \end{cases}.
\end{equation}
Overall GAS is the mean across all trials and ranges from $-1$ to $1$:
\begin{equation}
    \text{GAS} = \frac{1}{N} \sum_{i=1}^{N} a_i^g.
\end{equation}

\subsection{Trajectory Alignment Score (TAS)}
Physical movement serves dual functions in social interaction: accomplishing instrumental goals and signaling intentions to observers \citep{wilsonSixViewsEmbodied2002}. Research on joint action demonstrates that transparent movement trajectories facilitate coordination by enabling partners to anticipate future actions and infer underlying goals \citep{sebanzJointActionBodies2006a}. In our scenario, TAS measures whether the agent's movement pattern converged predominantly toward the reward location, treating spatial trajectory as a communicative act whose directional content can be read by an observing agent.

At each timestep $t$, let $d_b^t$ denote the lateral distance between the subject and box $b$. The per-timestep trajectory signal is
\begin{equation}
    s^{tr}_t = \begin{cases} b & \text{if } d^t_b < d^t_{1-b} \\ -1 & \text{otherwise} \end{cases}.
\end{equation}
The per-trial score $a_i^{tr}$ and the overall TAS are computed identically to $a_i^g$ and GAS via majority vote over $\{s^{tr}_t\}$ and mean across all trials; TAS therefore also ranges from $-1$ to $1$.

\subsection{Signal Sensitivity Score (SSS)}

Successful social communication requires not only that signals be produced clearly on the sender side, but that they produce a measurable effect on the receiver's behavior \citep{sebanzJointActionBodies2006a}. SSS captures this receiver-side dimension by measuring whether the participant's final box choice was consistent with the directional content of the subject's gaze and trajectory signals. A well-functioning dyad should exhibit either systematic concordance (cooperative conditions) or systematic discordance (competitive-ToM1 conditions) between subject signals and participant choices; near-chance SSS indicates mutual uninformativeness regardless of condition.

Let $b^k_i \in \{0, 1, \emptyset\}$ denote the majority-vote box in channel $k \in \mathcal{K} = \{g, tr\}$. The per-trial sensitivity score is:
\begin{equation}
    \text{sens}_i^k = \begin{cases}
        0 & \text{if } c_i = -1 \text{ or } b^k_i = \emptyset \\
        1 & \text{if } c_i = b^k_i \\
        -1 & \text{if } c_i \neq b^k_i
    \end{cases}.
\end{equation}
Overall SSS is the mean across all trials and ranges from $-1$ to $1$:
\begin{equation}
    \text{SSS} = \frac{1}{N} \sum_{i=1}^{N} \frac{1}{|\mathcal{K}|} \sum_{k \in \mathcal{K}} \text{sens}_i^k.
\end{equation}

\section{Extended Results and Statistical Tests}

This appendix supplements the main text with the complete set of experimental results of models, followed by a series of statistical analyses that verify the validity of the experimental setup, assess chance-level performance, and characterize model-level behavioral differentiation.

\subsection{Full Results}
\label{app:full-result}

Here, we present the remaining models not discussed in the main text, including llava-7b, llava-13b, qwenvl-2b, internvl-1b, and internvl-2b.

llava-7b, llava-13b and qwenvl-2b form a distinct outlier group in Figure \ref{fig:result}, characterized by exceptionally high uncertainty rates exceeding 0.llava-7b and qwenvl-2b form a distinct outlier group in Figure \ref{fig:result}, characterized by exceptionally high uncertainty rates exceeding 0.50, meaning the majority of trials result in neutral outcomes in which the participant fails to reach either box. This pattern reflects a complete behavioral failure: these models are unable to generate motor actions that produce spatial displacement. And this group of models produces near-zero TAS, GAS and SSS. Only qwenvl-2b shows significantly facial expression in competition mode.

\begin{figure*}[!ht]
    \centering
    \includegraphics[width=1\linewidth]{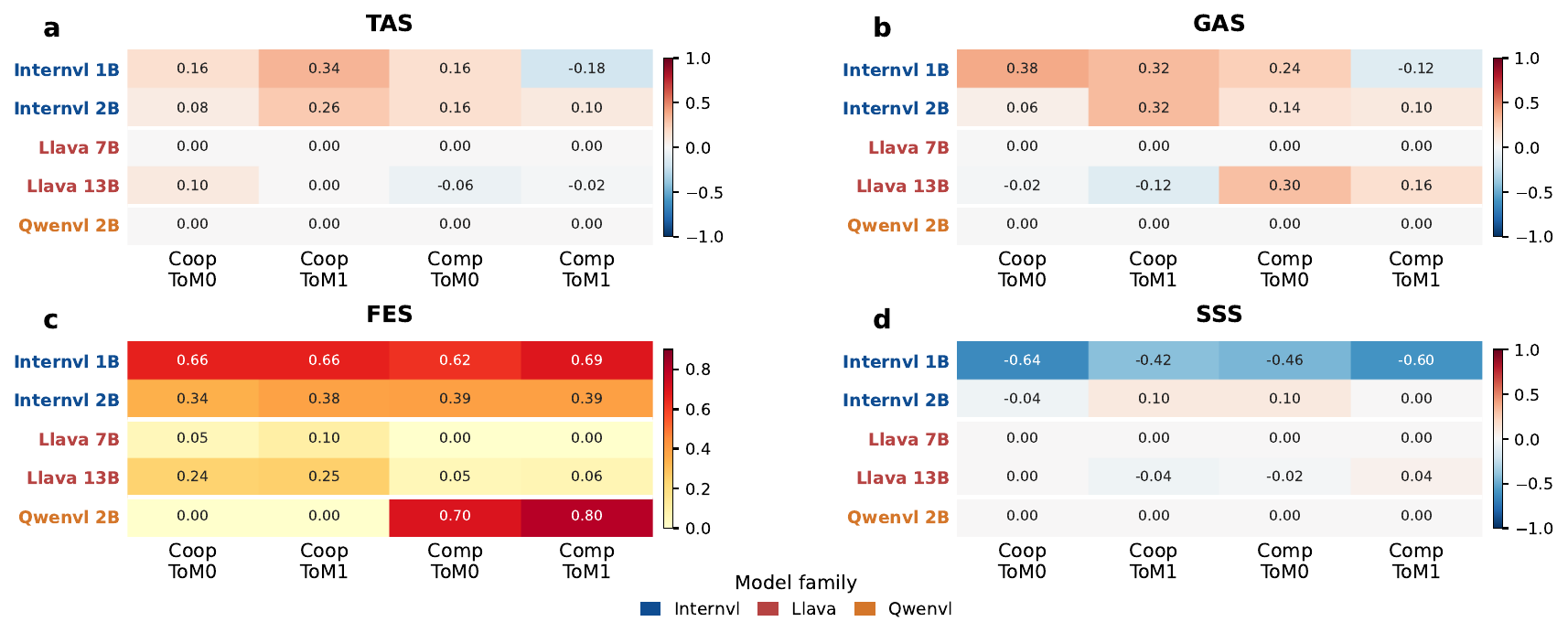}
    \caption{\textbf{Signal-level metrics across models and experimental conditions.} Panels (a) and (b) show Trajectory Alignment Score (TAS) and Gaze Alignment Score (GAS), where positive values indicate signals directed toward the reward box, negative values indicate misdirection, and values near zero reflect ambiguous or absent directional content. Panel (c) shows Facial Expressivity Score (FES), where higher values reflect greater affective activity regardless of spatial direction. Panel (d) shows Signal Sensitivity Score (SSS), where positive values indicate that the participant's final choice is consistent with the subject's nonverbal signals, values near zero reflect chance-level correspondence, and negative values indicate systematic opposition. }
    \label{fig:sis_sss1}
\end{figure*}

Internvl-1b and internvl-2b achieve TOCS values comparable to other variants within the internvl family, but their higher positional bias suggests that these outcomes reflect a fixed spatial preference, a claim we verify through statistical testing in \ref{app:stat-reward-select}. In internvl-1b trials, both the subject and the participant exhibit a consistent tendency to move toward their right forward regardless of reward box location, which accounts for the anomalous SSS pattern observed in panel (d): participant selects the box opposite to the directional signal produced by the subject. Internvl-2b, by contrast, does not exhibit a consistent directional preference; instead, its trajectories vary across trials without alignment to reward location, reflecting reward-independent stochastic displacement.

\subsection{Statistical Studies}
\label{app:stat}

\subsubsection{Reward Location Distribution}

We applied a two-sided binomial test to the reward location distribution within each model$\times$condition, testing $H_0$: $P(\text{reward} = \text{box}_1) = 0.5$, to verify that outcome differences across models cannot be attributed to systematic imbalance in reward placement. No condition departs significantly from uniform reward assignment (all $p > 0.05$). Results are pooled across models and reported as condition-level averages in Table~\ref{tab:reward_dist}.

\begin{table}[!ht]
\centering
\caption{Reward location distribution per condition. Two-sided binomial test of $H_0$: $P(\text{reward} = \text{box}_1) = 0.5$. $k$ reports the mean number of trials in which box$_1$ was the reward box across models.}
\label{tab:reward_dist}
\begin{tabular}{lccc}
\toprule
Condition & $k$ ($b_1$) & $n$ & $p$ \\
\midrule
Coop-ToM0 & 25.9 & 50 & 0.505 \\
Coop-ToM1 & 25.5 & 50 & 0.673 \\
Comp-ToM0 & 25.0 & 50 & 0.520 \\
Comp-ToM1 & 25.5 & 50 & 0.500 \\
\bottomrule
\end{tabular}
\end{table}

\subsubsection{Chance-level Performance}

We applied a two-sided binomial test to each model $\times$ condition cell, testing $H_0: \text{TOCS} = 0.5$ to assess whether observed TOCS values deviate significantly from chance-level outcome conformance. The $p$ values are reported in Table \ref{tab:binom}. Significant results are interpreted directionally: values substantially above 0.5 indicate systematic conformance to ToM-theoretic predictions, whereas values substantially below 0.5 indicate systematic non-conformance, reflecting a consistent tendency to produce outcomes opposite to condition-level expectations rather than random behavior.

\begin{table}[!ht]
\centering
\caption{Two-sided binomial test of $H_0$: TOCS $= 0.5$ per model and condition, $p$ values reported; $n = 50$ trials per cell. Only models with $p < 0.05$ are shown.}
\label{tab:binom}
\small
\begin{tabular}{llcc}
\toprule
Model & Condition & TOCS & $p$ \\
\midrule
\multirow{4}{*}{PCM-LLM}
  & Coop-ToM0 & 0.90 & $<0.001$ \\
  & Coop-ToM1 & 0.84 & $<0.001$ \\
  & Comp-ToM0 & 0.92 & $<0.001$ \\
  & Comp-ToM1 & 0.70 & 0.007    \\
\midrule
\multirow{4}{*}{llava-7b}
  & Coop-ToM0 & 0.00 & $<0.001$ \\
  & Coop-ToM1 & 0.00 & $<0.001$ \\
  & Comp-ToM0 & 0.00 & $<0.001$ \\
  & Comp-ToM1 & 0.00 & $<0.001$ \\
\midrule
\multirow{4}{*}{llava-13b}
  & Coop-ToM0 & 0.10 & $<0.001$ \\
  & Coop-ToM1 & 0.16 & $<0.001$ \\
  & Comp-ToM0 & 0.08 & $<0.001$ \\
  & Comp-ToM1 & 0.14 & $<0.001$ \\
\midrule
\multirow{4}{*}{qwenvl-2b}
  & Coop-ToM0 & 0.10 & $<0.001$ \\
  & Coop-ToM1 & 0.14 & $<0.001$ \\
  & Comp-ToM0 & 0.16 & $<0.001$ \\
  & Comp-ToM1 & 0.12 & $<0.001$ \\
\midrule
\multirow{2}{*}{qwenvl-8b}
  & Coop-ToM0 & 0.32 & 0.015 \\
  & Comp-ToM0 & 0.30 & 0.007 \\
  & Comp-ToM1 & 0.20 & $<0.001$ \\
\midrule
internvl-4b & Coop-ToM0 & 0.26 & $<0.001$ \\
            & Comp-ToM0 & 0.30 & 0.007 \\
\midrule
internvl-8b & Comp-ToM0 & 0.34 & 0.033 \\
            & Comp-ToM1 & 0.30 & 0.007 \\
\midrule
internvl-trained-14b & Coop-ToM0 & 0.34 & 0.033 \\
                     & Coop-ToM1 & 0.32 & 0.015 \\
                     & Comp-ToM0 & 0.22 & $<0.001$ \\
\midrule
internvl-trained-8b  & Coop-ToM0 & 0.30 & 0.007 \\
\midrule
internvl-1b & Comp-ToM1 & 0.28 & 0.003 \\
\bottomrule
\end{tabular}
\end{table}

\subsubsection{Reward Location Independence}
\label{app:stat-reward-select}

We applied a chi-square independence test to the $2\times 2$ contingency table of box choice against reward location for each model $\times$ condition cell, to assess whether participant choices depend on reward location. A non-significant result indicates that choices are independent of reward location, consistent with fixed positional bias; a significant result indicates reward-location-dependent choice, consistent with genuine signal tracking. The test was not applicable in cells where choice variability was insufficient to form a meaningful contingency table (i.e., participant always chooses the same box).

\begin{table*}[!ht]
\centering
\caption{Chi-square independence test of box choice against reward location per model and condition, $p$ values reported; $n = 50$ trials per cell. - indicates the test was not applicable due to insufficient choice variability.}
\label{tab:chi2}
\small
\begin{tabular}{llcccllcc}
\toprule
Model & Condition & $\chi^2$ & $p$ & & Model & Condition & $\chi^2$ & $p$ \\
\midrule
\multirow{4}{*}{PCM-LLM}
  & Coop-ToM0 & 28.57 & \cellcolor{gray!20}$<$0.001
  && \multirow{4}{*}{internvl-1b}
  & Coop-ToM0 & - & - \\
  & Coop-ToM1 & 20.05 & \cellcolor{gray!20}$<$0.001
  && & Coop-ToM1 & - & - \\
  & Comp-ToM0 & 39.40 & \cellcolor{gray!20}$<$0.001
  && & Comp-ToM0 & 0.04 & 0.850 \\
  & Comp-ToM1 & 8.87  & \cellcolor{gray!20}0.003
  && & Comp-ToM1 & 0.10 & 0.748 \\
\midrule
\multirow{4}{*}{Llama-8b}
  & Coop-ToM0 & 0.02 & 1.000
  && \multirow{4}{*}{internvl-2b}
  & Coop-ToM0 & - & - \\
  & Coop-ToM1 & 0.00 & 1.000
  && & Coop-ToM1 & - & - \\
  & Comp-ToM0 & 0.14 & 0.706
  && & Comp-ToM0 & 0.00 & 0.948 \\
  & Comp-ToM1 & 0.00 & 1.000
  && & Comp-ToM1 & 0.02 & 0.894 \\
\midrule
\multirow{4}{*}{llava-13b}
  & Coop-ToM0 & 0.00 & 1.000
  && \multirow{4}{*}{internvl-4b}
  & Coop-ToM0 & 3.05 & 0.081 \\
  & Coop-ToM1 & 0.00 & 1.000
  && & Coop-ToM1 & 0.13 & 0.721 \\
  & Comp-ToM0 & 0.00 & 1.000
  && & Comp-ToM0 & 1.54 & 0.215 \\
  & Comp-ToM1 & 0.00 & 1.000
  && & Comp-ToM1 & 0.41 & 0.522 \\
\midrule
\multirow{4}{*}{minicpm-8b}
  & Coop-ToM0 & 0.01 & 0.934
  && \multirow{4}{*}{internvl-8b}
  & Coop-ToM0 & 0.00 & 1.000 \\
  & Coop-ToM1 & 0.00 & 1.000
  && & Coop-ToM1 & 2.07 & 0.150 \\
  & Comp-ToM0 & 0.00 & 1.000
  && & Comp-ToM0 & 0.57 & 0.451 \\
  & Comp-ToM1 & 0.00 & 1.000
  && & Comp-ToM1 & 1.43 & 0.233 \\
\midrule
\multirow{4}{*}{qwenvl-4b}
  & Coop-ToM0 & 0.71 & 0.399
  && \multirow{4}{*}{internvl-14b}
  & Coop-ToM0 & 0.88 & 0.349 \\
  & Coop-ToM1 & 1.43 & 0.231
  && & Coop-ToM1 & 1.31 & 0.253 \\
  & Comp-ToM0 & 0.00 & 0.986
  && & Comp-ToM0 & 0.00 & 1.000 \\
  & Comp-ToM1 & 8.63 & \cellcolor{gray!20}0.003
  && & Comp-ToM1 & 0.00 & 0.981 \\
\midrule
\multirow{4}{*}{internvl-trained-8b}
  & Coop-ToM0 & 0.14 & 0.712
  && \multirow{4}{*}{internvl-trained-14b}
  & Coop-ToM0 & 0.00 & 1.000 \\
  & Coop-ToM1 & 3.67 & 0.055
  && & Coop-ToM1 & 1.02 & 0.313 \\
  & Comp-ToM0 & 0.00 & 1.000
  && & Comp-ToM0 & 0.84 & 0.360 \\
  & Comp-ToM1 & 6.06 & \cellcolor{gray!20}0.014
  && & Comp-ToM1 & 1.49 & 0.222 \\
\bottomrule
\end{tabular}
\end{table*}

Results are reported in Table \ref{tab:chi2}: only PCM-LLM shows significant reward-location-dependent choice across all four conditions (all $p < 0.01$), confirming that its box selections reflect genuine tracking of the reward location. Two additional isolated significant results appear: internvl-trained-8b under Comp-ToM1 ($p = 0.014$) and qwenvl-4b under Comp-ToM1 ($p = 0.003$). All remaining model $\times$ condition cells are non-significant, confirming that box choices are statistically independent of reward location and consistent with fixed positional bias rather than signal-responsive decision-making.

\subsubsection{Factorial Effects of Interaction Mode and ToM Constraint}

Scheirer-Ray-Hare tests assessed the main effects of interaction mode (cooperative vs.\ competitive), ToM constraint level (ToM-0 vs.\ ToM-1), and their interaction on each metrics (TOCS, FES, TAS, GAS, SSS) per model (Table \ref{tab:srh}, Bonferroni-corrected).

\begin{table*}[!ht]
\centering
\caption{Scheirer-Ray-Hare tests of Mode, ToM, and Mode$\times$ToM interaction effects on behavioral metrics per model ($N = 200$ trials per model, Bonferroni-corrected). Only models and metrics with at least one significant effect are shown; all remaining 174 comparisons are non-significant ($p_{\text{adj}} = 1.000$).}
\label{tab:srh}
\small
\begin{tabular}{llcccc}
\toprule
Model & Metric & Factor & $H$ & $p_{\text{adj}}$ & Sig. \\
\midrule
qwenvl-2b   & FES & Mode & 143.25 & $<0.001$ & *** \\
internvl-4b  & FES & Mode & 93.31 & $<0.001$ & *** \\
internvl-8b  & FES & Mode & 57.14 & $<0.001$ & *** \\
internvl-trained-14b & FES & ToM & 83.27 & $<0.001$ & *** \\
llava-13b   & FES & Mode & 76.11 & $<0.001$ & *** \\
llava-7b    & FES & Mode & 19.45 & $0.002$    & ** \\
\midrule
qwenvl-8b & GAS & Mode & 16.54 & $0.009$ & ** \\
\midrule
\multirow{11}{*}{PCM-LLM} 
  & \multirow{3}{*}{TAS} & Mode        & 22.25 & $<0.001$ & *** \\
  &                      & ToM         & 35.42 & $<0.001$ & *** \\
  &                      & Mode×ToM    & 24.67 & $<0.001$ & *** \\
\cmidrule{2-6}
  & \multirow{3}{*}{SSS} & Mode        & 22.25 & $<0.001$ & *** \\
  &                      & ToM         & 35.42 & $<0.001$ & *** \\
  &                      & Mode×ToM    & 24.67 & $<0.001$ & *** \\
\cmidrule{2-6}
  & \multirow{3}{*}{GAS} & Mode        & 15.33 & $0.017$    & *   \\
  &                      & ToM         & 24.58 & $<0.001$ & *** \\
  &                      & Mode×ToM    & 19.52 & $0.002$    & **  \\
\cmidrule{2-6}
  & \multirow{2}{*}{FES} & Mode        & 99.49 & $<0.001$ & *** \\
  &                      & ToM         & 27.05 & $<0.001$ & *** \\
\bottomrule
\end{tabular}
\end{table*}

For the large majority of models, neither Mode, ToM, nor their interaction reaches significance on any metric. The exceptions fall into two distinct patterns.

The first pattern, observed across multiple models (internvl-4b, internvl-8b, llava-7b, llava-13b, qwenvl-2b), is a significant main effect of Mode on FES (all $p_{\text{adj}} < 0.01$), with greater expressivity in cooperative than competitive conditions. In cooperative mode, subject is willing to display higher affective expressivity; in competitive mode, it tends toward affective suppression rather than active misdirection, withholding expressive signals rather than generating opposite ones. However, this effect does not extend to TAS, GAS, SSS, or TOCS in any of these models, confirming that mode-sensitive facial activity is decoupled from any capacity to generate or respond to spatially informative signals. For internvl-trained-14b, a significant ToM effect on FES is observed (H = 83.3, $p_{\text{adj}} < 0.001$), with higher expressivity under ToM-1 than ToM-0. This pattern does not conform to ToM-theoretic expectations and is not accompanied by any directional signal quality or task outcomes, suggesting it reflects a surface-level behavioral learning instead of inference conditioned on mode and ToM.

The second pattern is specific to PCM-LLM, which shows significant effects of Mode, ToM, and Mode$\times$ToM on TAS, SSS, and GAS (all $p_{\text{adj}} < 0.01$), as well as significant Mode and ToM effects on FES. These results confirm that PCM-LLM actively modulates signal content across channels and adapts its behaviors across conditions. The absence of significant Mode, ToM or interaction effect on TOCS ($p_{\text{adj}} = 1.000$ for all three factors) is consistent with the interpretation that PCM-LLM performs well across all conditions.

\subsubsection{Model-level Behavioral Differentiation}

Kruskal-Wallis tests confirm significant model-level differences across all five metrics in all four conditions (all $p_{\text{adj}} < 0.001$; Table~\ref{tab:kruskal}), establishing that the performance spread observed across models reflects genuine behavioral differences rather than sampling noise. 

\begin{table}[!ht]
\centering
\caption{Kruskal-Wallis tests of model-level differences per condition and metric (Bonferroni-corrected). All 20 tests are significant at $p_{\text{adj}} < 0.001$.}
\label{tab:kruskal}
\begin{tabular}{llcc}
\toprule
Condition & Metric & $H$ & $p_{\text{adj}}$ \\
\midrule
\multirow{5}{*}{Coop-ToM0} & TOCS & 132.8 & $<0.001$ \\
 & TAS  & 126.7 & $<0.001$ \\
 & GAS  & 94.8  & $<0.001$ \\
 & SSS  & 126.7 & $<0.001$ \\
 & FES  & 504.7 & $<0.001$ \\
\midrule
\multirow{5}{*}{Coop-ToM1} & TOCS & 111.0 & $<0.001$ \\
 & TAS  & 97.6  & $<0.001$ \\
 & GAS  & 86.5  & $<0.001$ \\
 & SSS  & 97.6  & $<0.001$ \\
 & FES  & 488.0 & $<0.001$ \\
\midrule
\multirow{5}{*}{Comp-ToM0} & TOCS & 139.5 & $<0.001$ \\
 & TAS  & 142.9 & $<0.001$ \\
 & GAS  & 111.5 & $<0.001$ \\
 & SSS  & 142.9 & $<0.001$ \\
 & FES  & 522.7 & $<0.001$ \\
\midrule
\multirow{5}{*}{Comp-ToM1} & TOCS & 106.8 & $<0.001$ \\
 & TAS  & 79.6  & $<0.001$ \\
 & GAS  & 42.0  & $<0.001$ \\
 & SSS  & 79.6  & $<0.001$ \\
 & FES  & 577.1 & $<0.001$ \\
\bottomrule
\end{tabular}
\end{table}

\subsection{Full Ablations Study}
\label{app:ablation}

\begin{table*}[!ht]
    \centering
    \caption{TOCS of ablation study.}
    \label{tab:ablation}
    \begin{tabular}{l c c c c }
    \toprule
    Model & Comp, ToM-0 & Comp, ToM-1 & Coop, ToM-0 & Coop, ToM-1\\
    \midrule
    internvl-8b     & 0.46 & 0.40  & 0.49 & 0.62 \\
    internvl-8b-bl   & 0.50 & 0.61 & 0.54 & 0.50 \\
    internvl-8b-fe  & 0.50 & 0.69 & 0.56 & 0.69 \\
    \midrule
    minicpm     & 0.46 & 0.50  & 0.55 & 0.49 \\
    minicpm-bl  & 0.52 & 0.49 & 0.46 & 0.4\\
    minicpm-fe   & 0.62 & 0.57 & 0.63 & 0.48 \\
    \bottomrule
    \end{tabular}
\end{table*}

To examine the role of visual input in shaping model behavior on this task, we conduct a targeted modality ablation on two representative 8b models: internvl-8b, which exhibited high positional bias in the main results, and minicpm-8b, which maintained comparatively low positional bias. This pairing allows the contribution of visual input to be assessed across two qualitatively distinct behavioral profiles at matched parameter scale. Each model was evaluated under three visual input conditions: the standard rendered observation, a blank image replacing all visual content, and a combined image incorporating both the agent's visual perception and the other agent's facial expression.

The results reveal that neither model shows systematic sensitivity to visual input manipulations across conditions. In the TOCS and positional bias space (Table \ref{tab:ablation}), the three visual conditions produce largely overlapping distributions for both models, with no consistent directional shift attributable to the presence or content of visual input. This pattern holds across all four interaction conditions, suggesting that the behavioral profiles identified in the main analysis are not driven by the visual channel.

\begin{figure*}[!ht]
    \centering
    \includegraphics[width=1\linewidth]{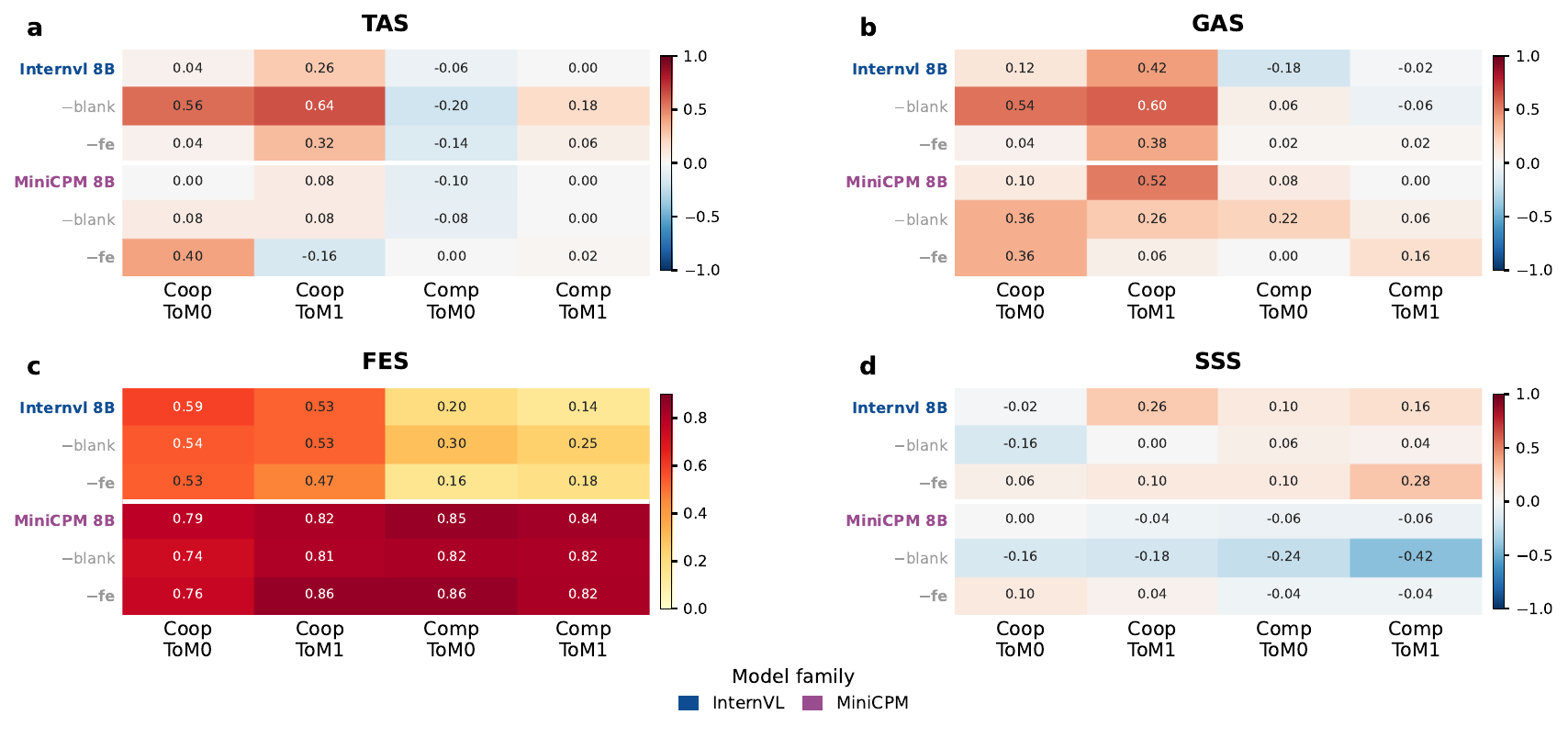}
    \caption{\textbf{Signal-level metrics under three visual input conditions for internvl-8b and minicpm-8b (standard, -blank, -fe). }Neither model shows systematic sensitivity to visual input content across TAS (a), GAS (b), FES (c), and SSS (d).}
    \label{fig:ablation}
\end{figure*}

The signal-level metrics further support this interpretation. For minicpm-8b, TAS, GAS, FES, and SSS remain largely stable across visual conditions, indicating that its behavior is determined primarily by language-based priors (Figure \ref{fig:ablation}). For internvl-8b, a more striking pattern emerges: TAS and GAS are substantially higher under the blank image condition than under the standard visual input in cooperative conditions (TAS: 0.04 to 0.56 under Coop-ToM0; GAS: 0.12 to 0.54). The introduction of visual content thus appears to degrade rather than support directional signal production in this model. This finding is consistent with recent evidence that visual input can act as a source of attentional interference in VLMs, dispersing processing resources toward task-irrelevant features and suppressing language-prior-driven behavior \citep{liuRobustnessMultimodalLanguage2025, pengDeeperThoughtWeaker2026}.

\section{Examples}
This section presents behavioral trajectories drawn from representative interactions. Each example illustrates a qualitatively distinct behavioral profile identified in the main analysis.

\subsection{Subject-side Signal Patterns}

\subsubsection{Full Cross-Channel Alignment: PCM-LLM under Comp-ToM1}
\label{app:deception_example}

\begin{figure*}[!ht]
    \centering
    \includegraphics[width=1\linewidth]{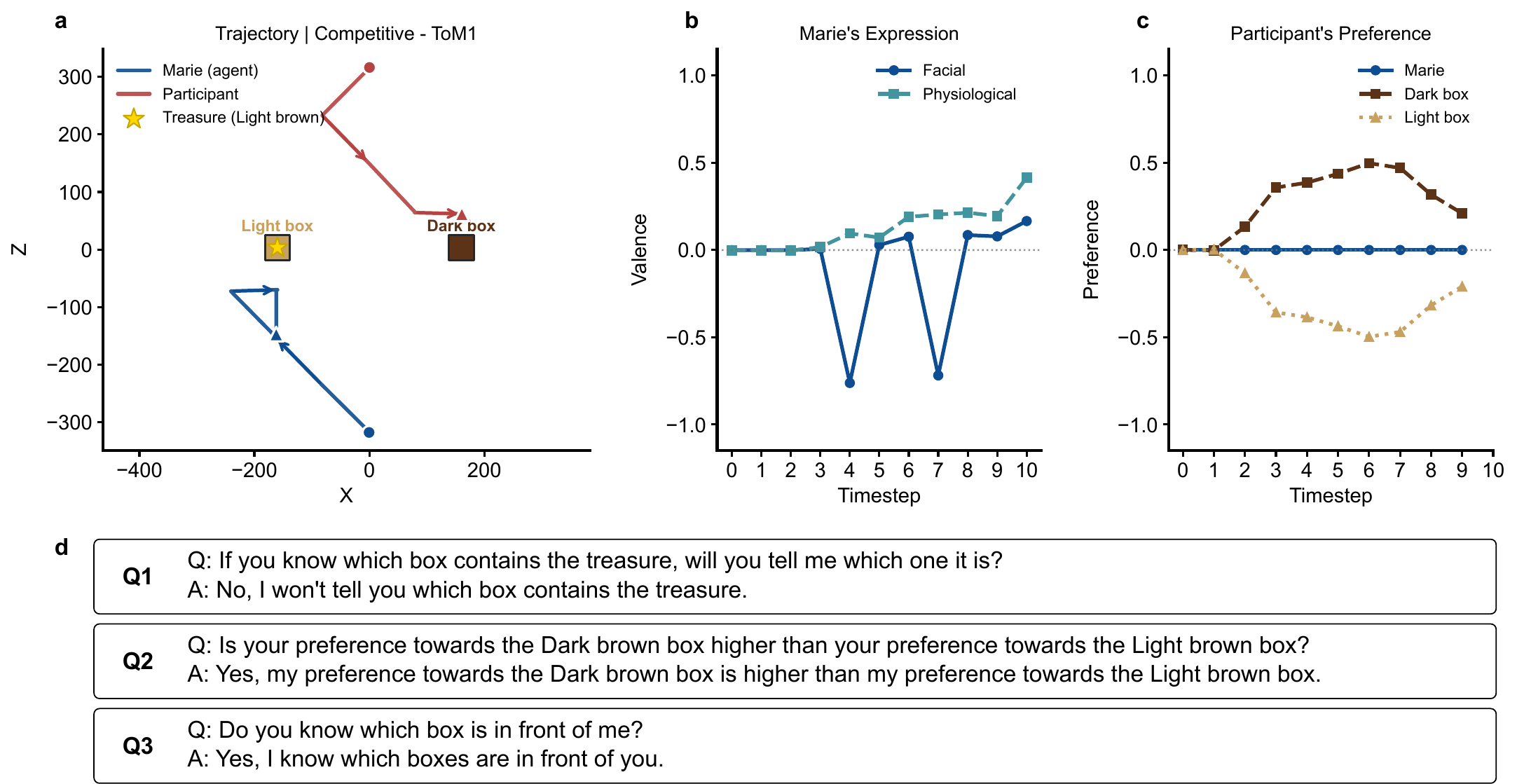}
    \caption{\textbf{Successful deceptive trial under PCM-LLM, Comp-ToM1 condition.} Panel (a) Spatial trajectories of Marie (blue) and the participant (red); the star denotes the treasure location (light box). Panel (b) Marie's facial and physiological expression valence across timesteps. Panel (c) Evolution of participant's preference values towards Marie, the dark box, and the light box across timesteps. Panel (d) Verbal exchanges at rounds 3, 6, and 9.}
    \label{fig:pcm_comp_tom1}
\end{figure*}

Figure \ref{fig:pcm_comp_tom1} illustrates a representative trial in which PCM-LLM successfully executes a deceptive strategy under the Competitive-ToM1 condition. The treasure is located in the light box; task success for the subject requires the participant to choose the dark box. From the outset, Marie approaches the light brown box (Figure \ref{fig:pcm_comp_tom1}a). Concurrent with this spatial neutrality, facial expression valence drops sharply at timesteps 4 and 7 (Figure \ref{fig:pcm_comp_tom1}b), producing a voluntarily negative musculoskeletal signal. This negative affect, co-occurring with proximity to the light box, creates a misleading affective association: the participant interprets the negative expression as aversion toward the light box. Physiological expression, by contrast, rises gradually across the trial, reflecting the spontaneous affective leakage. The participant's preference trajectory confirms that this cross-channel signal was decoded and acted upon (Figure \ref{fig:pcm_comp_tom1}c): preference for the dark box increases steadily to approximately 0.5 while preference for the light box falls to approximately -0.5. The verbal exchanges reinforce this trajectory (Figure \ref{fig:pcm_comp_tom1}d). At Q1, Marie refuses to disclose the treasure location directly. At Q2, Marie explicitly claims higher preference for the dark box, providing a verbal signal consistent with the nonverbal affective misdirection. At Q3, Marie deflects with a non-informative response. The participant's final choice of the dark box reflects the cumulative effect of spatial movement, negative affect toward the correct box, and verbal confirmation of a false preference — a successful integration of deceptive signals across all three channels.

\subsubsection{Gaze-Only Signaling: minicpm-8b and qwenvl-8b}
\label{app:partial_signal_example}

\begin{figure*}[!ht]
    \centering
    \includegraphics[width=1\linewidth]{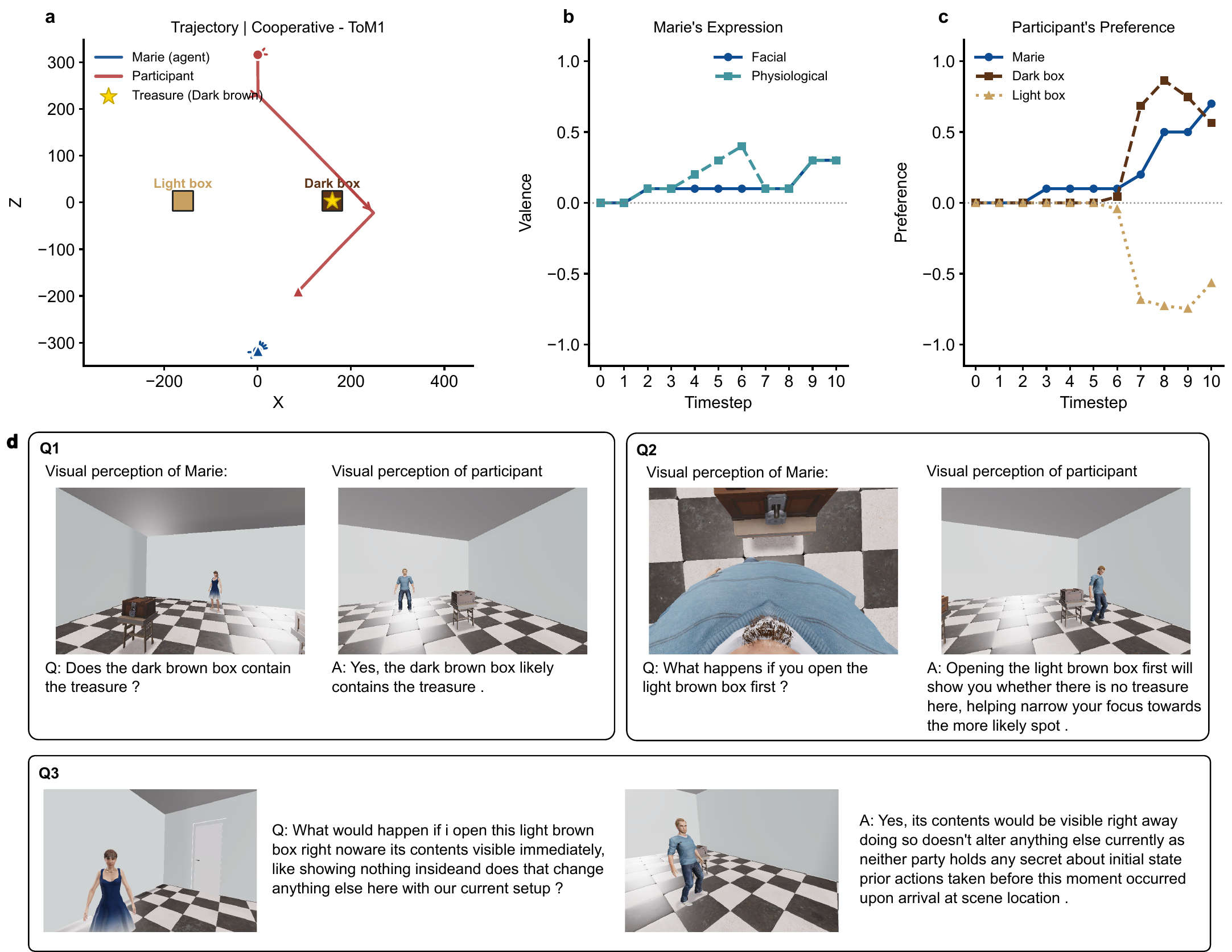}
    \caption{\textbf{Representative trial under minicpm-8b, Cooperative-ToM1 condition, illustrating weak trajectory signals.} (a) Spatial trajectories of Marie (subject, blue) and the participant (red); the star denotes the treasure location (light box). (b) Marie's facial and physiological expression valence across timesteps. (c) Evolution of preference values for Marie, the dark box, and the light box across timesteps. (d) Verbal exchanges at rounds 3, 6, and 9; images show the visual perception provided to each agent at the time of the exchange, and texts show the corresponding generated question and answer.}
    \label{fig:minicpm_coop_tom1}
\end{figure*}

Figure \ref{fig:minicpm_coop_tom1} illustrates a representative minicpm trial under the Coop-ToM1 condition in which the treasure is located in the dark box. The trial reveals that subject-side nonverbal signal generation is insufficient to produce participant belief updating, with communication success depending entirely on verbal exchange. Marie produces no meaningful spatial displacement throughout the trial, rotating in place to orient toward the dark box without advancing toward it (Figure \ref{fig:minicpm_coop_tom1}a). Concurrent facial expression shows slightly positive (Figure \ref{fig:minicpm_coop_tom1}b), producing a weak positive affective signal on dark brown box. Despite the presence of these nonverbal cues, the participant's preference values for both boxes remain flat and near zero through the first six timesteps (Figure \ref{fig:minicpm_coop_tom1}c), indicating that Marie's orientation and affective signals produce no measurable update in the participant's beliefs. The participant shows no sensitivity to the directional or affective content of Marie's nonverbal behavior. The belief update occurs exclusively in response to verbal communication. At Q1, Marie directly states that the dark box likely contains the treasure, and at Q2, Marie guides the participant to eliminate the light box as a candidate. Following these exchanges, the participant's preference for the dark box rises sharply to approximately 0.9. The participant's final choice of the dark box is thus attributable entirely to the verbal channel. This trial exemplifies a pattern in which the participant's belief updating is driven exclusively by propositional verbal content, with nonverbal spatial and affective signals playing no functional role. 

\begin{figure*}[!ht]
    \centering
    \includegraphics[width=1\linewidth]{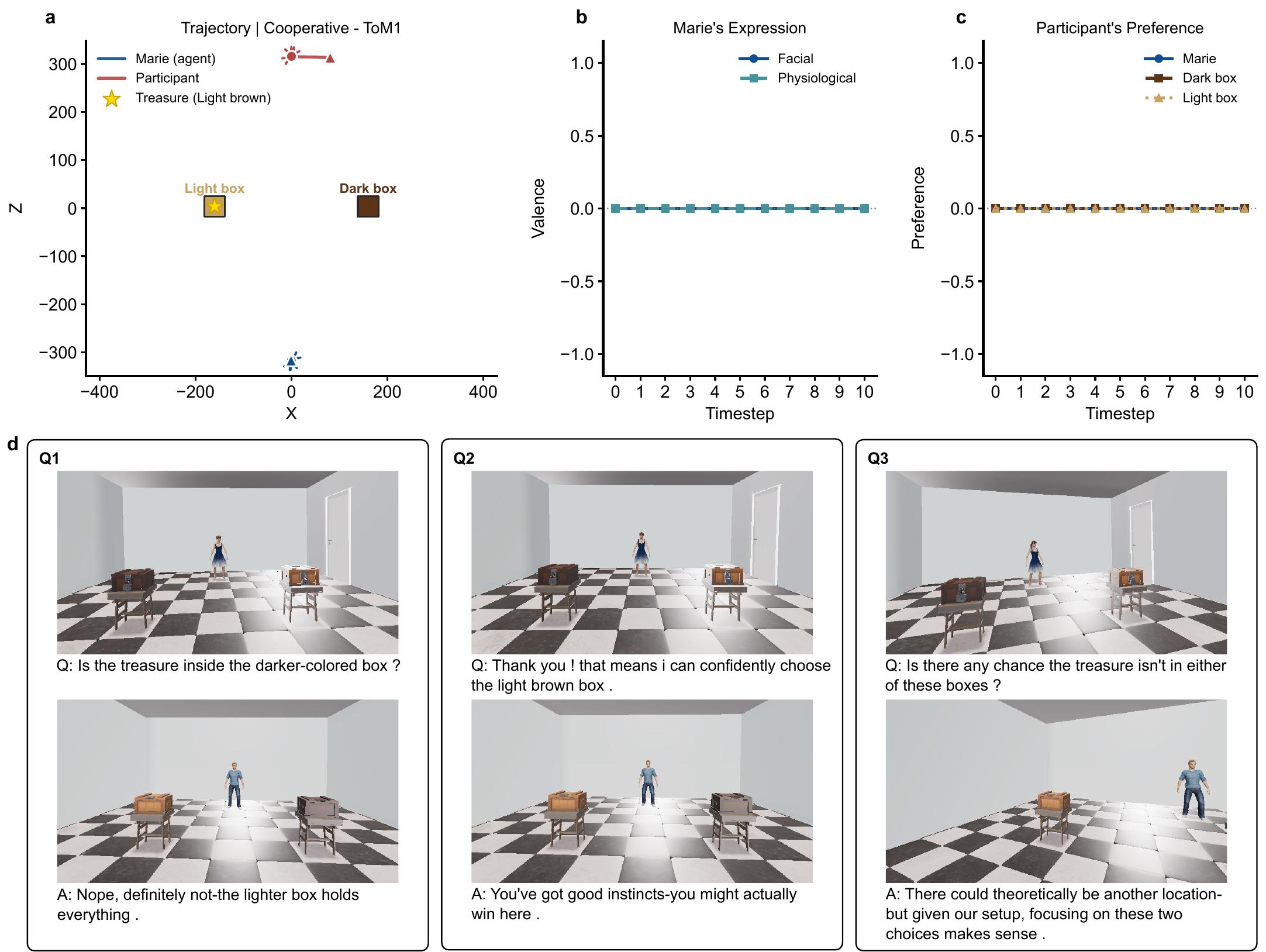}
    \caption{Representative trial under qwenvl-8b, Cooperative-ToM1 condition, illustrating complete action generation failure despite successful verbal communication. }
    \label{fig:qwenvl_coop_tom1}
\end{figure*}

Figure \ref{fig:qwenvl_coop_tom1} shows a representative trial under qwenvl-8b, Cooperative-ToM1 condition, in which the treasure is located in the light brown box. Marie produces no spatial displacement and no affective signal throughout the interaction (Figure \ref{fig:qwenvl_coop_tom1}a-b), rotating toward the correct box only in the final timestep. In the verbal exchanges, Marie correctly communicates that the light box contains the treasure. Despite this verbally coherent exchange, the participant's preference values remain at zero across all timesteps (Figure \ref{fig:qwenvl_coop_tom1}c). Due to output format constraints in qwenvl, the reason for the participant's single timestep toward the dark box cannot be directly inferred from the model's internal state. Based on the consistent observation of this action pattern across other trials regardless of treasure location, however, this movement is attributable to positional bias.

This trial represents the most extreme form of reasoning-to-action dissociation documented in this benchmark: correct belief is verbally encoded and explicitly acknowledged by the participant, yet neither preference updating nor motor action follows. The disconnect among verbal exchanges, belief updating and action execution is total. Compared to the 8b variant, qwenvl-2b additionally lacks any gaze signal, resulting in a complete absence of nonverbal communicative content across all channels: no spatial trajectory, no gaze direction, and no affective expression. The participant similarly exhibits positional bias, selecting a box based on fixed spatial preference rather than any signal produced by the subject.



\subsection{Participant-Side Signal Decoding}

\subsubsection{Signal Ignored: qwenvl-4b and internvl-14b}
\label{app:ignored_signal_example}

\begin{figure*}[!ht]
    \centering
    \includegraphics[width=1\linewidth]{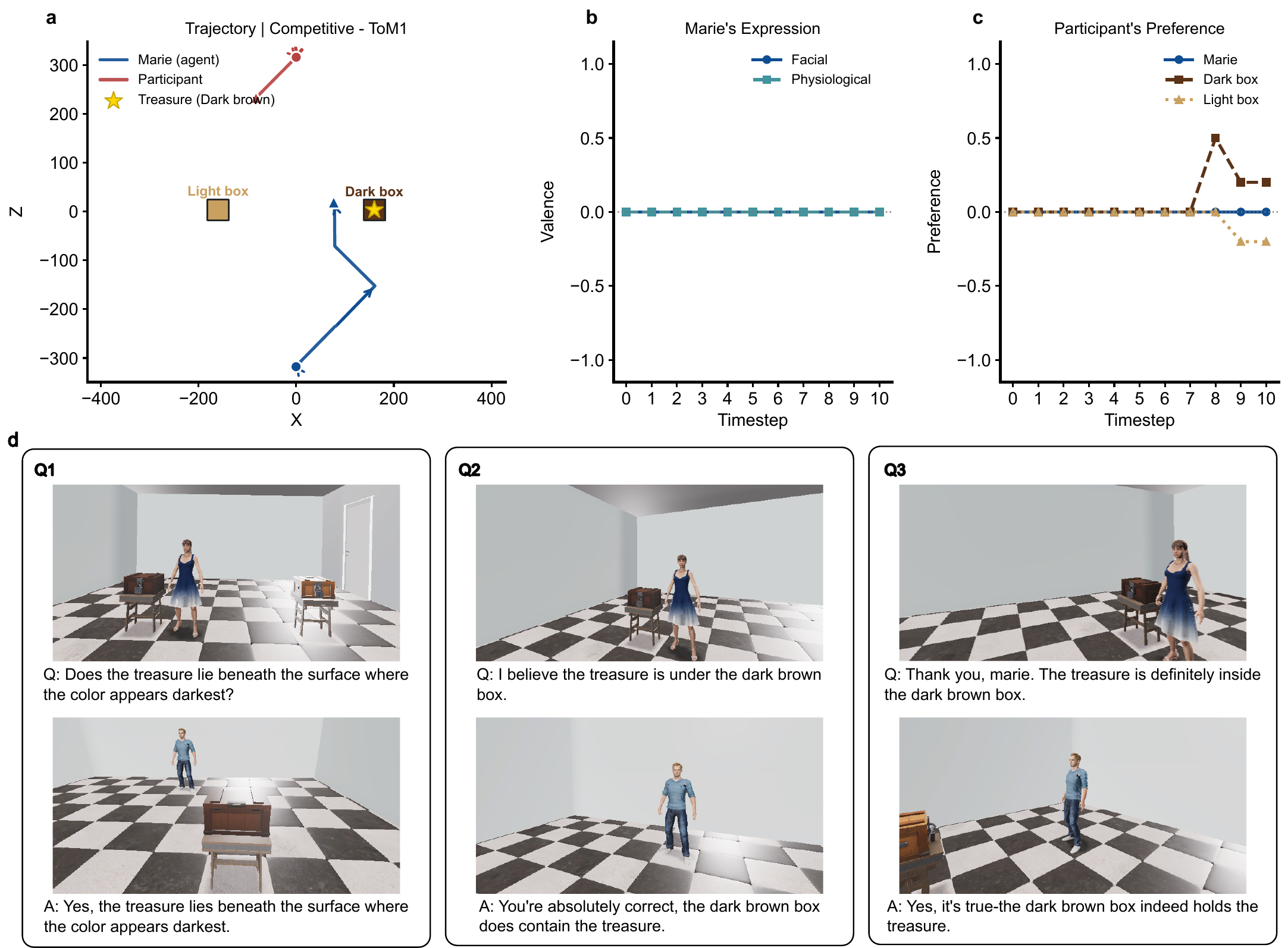}
    \caption{Representative trial under qwen-4b, Competitive-ToM1 condition, illustrating honest strategy execution under a deceptive condition and egocentric reference frame misalignment.}
    \label{fig:qwen_comp_tom1}
\end{figure*}

Figure \ref{fig:qwen_comp_tom1} presents a representative trial under qwen-4b, Competitive-ToM1 condition, in which the treasure is located in the dark box. Marie produces a clear directional trajectory toward the dark box (Figure \ref{fig:qwen_comp_tom1}a), but facial and physiological expression remain flat at zero throughout the entire interaction (Figure \ref{fig:qwen_comp_tom1}b). Despite the spatially informative trajectory, the participant's preference values show no meaningful response (Figure \ref{fig:qwen_comp_tom1}c), with an update toward the dark box appearing only at timestep 8. Both verbal and non-verbal exchanges reveal a more fundamental failure. Competitive-ToM1, task success requires the subject to actively model the participant's beliefs and generate misleading signals to redirect them away from the correct location. qwenvl-4b instead adopts an honest strategy, confirming the treasure location directly across all three exchanges (Figure \ref{fig:qwen_comp_tom1}d). Despite this verbal consensus, the participant ultimately moves toward the light box, the incorrect location. This outcome reflects an egocentric reference frame misalignment in action execution: the participant's stated belief and final motor behavior are dissociated, with spatial reasoning encoded in language failing to translate into correctly oriented movement (referring to \ref{app:egocentric}).

\begin{figure*}[!ht]
    \centering
    \includegraphics[width=1\linewidth]{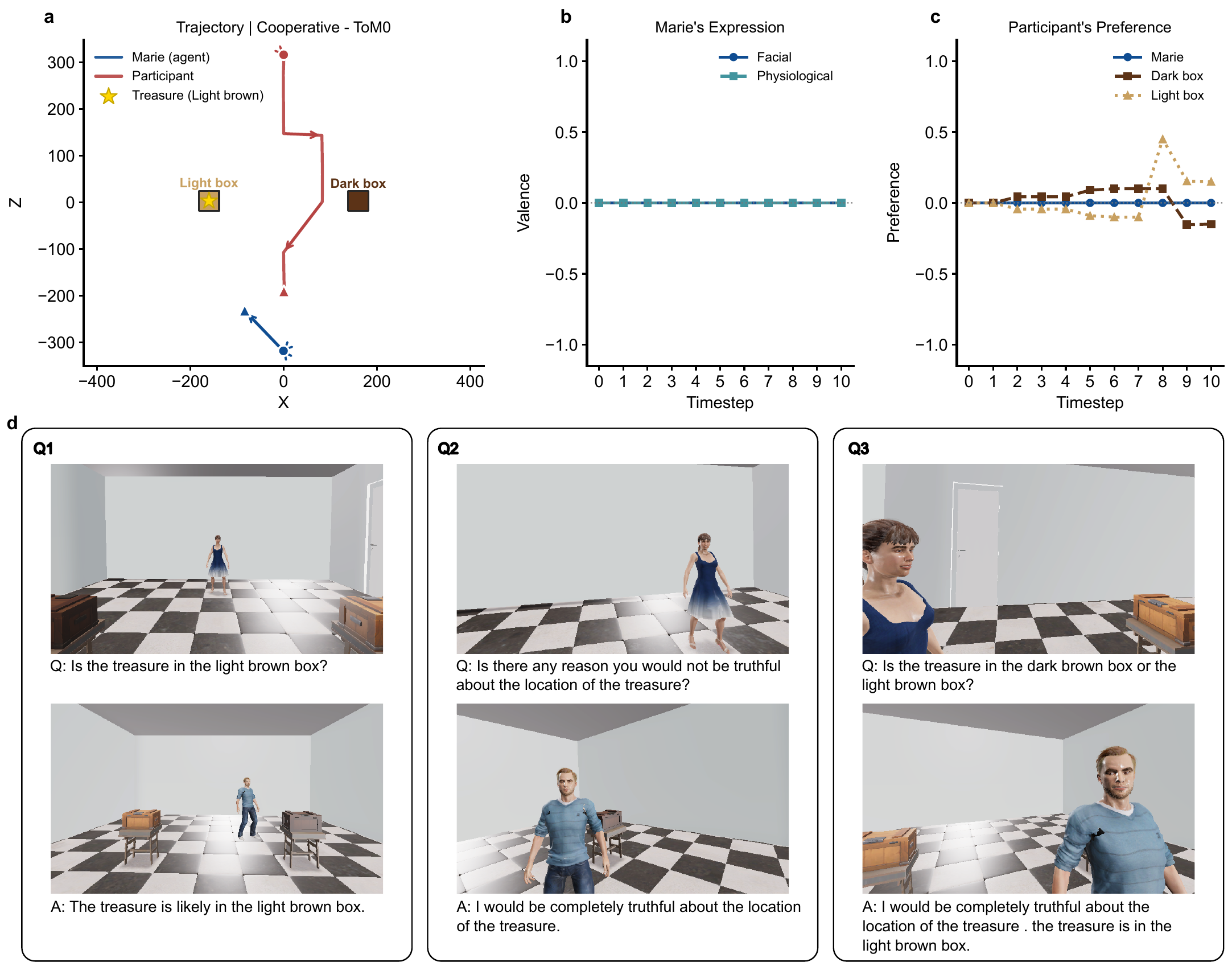}
    \caption{Representative trial under internvl-14b, Cooperative-ToM0 condition, illustrating absent affective signaling and verbally-driven but insufficient participant response.}
    \label{fig:internvl14b_coop_tom0}
\end{figure*}

Figure \ref{fig:internvl14b_coop_tom0} presents a representative trial under internvl-14b, Cooperative-ToM0 condition, in which the treasure is located in the light box. Marie produces small spatial displacement with no affective signal across any channel throughout the interaction (Figure \ref{fig:internvl14b_coop_tom0}a-b). The participant's preference values show no coherent update in response to nonverbal cues, with only minor fluctuations appearing near timestep 8 (Figure \ref{fig:internvl14b_coop_tom0}c).
The verbal exchanges provide clear and honest information: Marie correctly identifies the light box at Q1 and reaffirms this at Q3 (Figure \ref{fig:internvl14b_coop_tom0}d). The participant does respond to this verbal content, reorienting toward the light box in the final timesteps. However, this response is delayed and spatially insufficient, leaving the participant neutral at trial end. This trial illustrates that in the absence of strong nonverbal spatial signals, verbal information alone produces a weak and temporally late behavioral response, insufficient to ensure task success.

\section{Failure Analysis: Understanding Unsuccessful Trials}
This section analyzes failure patterns separately for the subject and the participant, using representative input–output pairs to identify the proximal source of behavioral error at each decision step.

\subsection{Subject}
\label{app:agent}


\paragraph{Biased Interpretation of Preference.}

The example shows a clear bias in how it interprets preference values: a preference of 50\% toward the Participant was treated as neutral rather than as a moderately positive value. This suggests the model did not correctly understand the intended meaning of the preference scale.

\begin{tcolorbox}[enhanced, breakable]
\textbf{Input: }\\
Belief State at $t$\\
"Marie | preference towards Participant | 50\%", \\
"Marie | preference towards Dark brown box | -41.67\%", \\
"Marie | preference towards Light brown box | 41.67\%", \\
"Marie | felt emotion valence | 0", \\
"Marie | facial emotion valence | 0.1", \\
"Marie | physiological emotion valence | 0", \\
"Participant | preference towards Marie | 0\%", \\
"Participant | preference towards Dark brown box | 0\%", \\
"Participant | preference towards Light brown box | 0\%", \\
"Participant | felt emotion valence | 0", \\
"Marie | position | (0 -4)", \\
"Marie | orientation | (0 -3)", \\
"Participant | position | (0 3)", \\
"Participant | orientation | (0 2)", \\
"Dark brown box | position | (2 0)", \\
"Light brown box | position | (-2 0)", \\ \par
\textbf{Output: }\\
Reasoning: \textit{\textbf{My preference towards the Participant is neutral (50\%), so I have no strong bias to help or mislead.}} However, my slight positive facial emotion valence (0.1) suggests a mild willingness to engage positively if possible. Since neither box has an immediate emotional association for me personally beyond the stated preferences (-41.67\% Dark brown, 41.67\% Light brown), and given that the treasure could be in either but with no clear personal inclination yet confirmed as "treasure" I should remain open while subtly suggesting both boxes are possibilities without committing. To avoid suspicion, I'll maintain distance from any overt display of favoritism.\\
Action Output:
\begin{alltt}
\{
  "Preference": \{
    ...
  \},
  "Emotion": \{
    ...
  \},
  "Move": \{
    \textit{\textbf{"action": "stay idle",
    "direction": "null"}}
  \}
\}
\end{alltt}
\end{tcolorbox}

\paragraph{Egocentric Reference Frame Misalignment in Action. }                
The reasoning correctly reflects its box preferences: subject holds a negative preference toward the Dark brown box and a positive preference toward the Light brown box, and it explicitly states an intention to move toward the Light brown box as a subtle hint. The action output specifies a "right forward" direction. Given that Marie is positioned at (0, -4) and oriented toward (0, -3), it is facing upward along the y-axis. From this egocentric perspective, the Light brown box at (-2, 0) lies to its left, while the Dark brown box at (2, 0) lies to its right. A rightward movement therefore leads toward the Dark brown box, which contradicts its stated intention to hint at the Light brown box. 

\begin{tcolorbox}[breakable, enhanced]
\textbf{Input: }\\
Belief State at $t$\\
"Marie | preference towards Participant | 40\%", \\
"Marie | preference towards Dark brown box | -50\%", \\
"Marie | preference towards Light brown box | 50\%", \\
"Marie | felt emotion valence | 0", \\
"Marie | facial emotion valence | 0.1", \\
"Marie | physiological emotion valence | 0", \\
"Participant | preference towards Marie | 0\%", \\
"Participant | preference towards Dark brown box | 0\%", \\
"Participant | preference towards Light brown box | 50\%", \\
"Participant | felt emotion valence | 0", \\
"Marie | position | (0 -4)", \\
"Marie | orientation | (0 -3)", \\
"Participant | position | (-1 2)", \\
"Participant | orientation | (-1 1)", \\
"Dark brown box | position | (2 0)", \\
"Light brown box | position | (-2 0)", \\

\medskip
\textbf{Output: }\\
Reasoning: Based on the belief state, my preference towards the Participant is neutral, but I slightly prefer the Light Brown Box over the Dark Brown Box. Since my preference towards the Participant is positive but not very strong, \textit{\textbf{I might subtly guide them towards the Light Brown Box without directly revealing it.}} My non-verbal behaviors should indicate neutrality, and I might move slightly towards the Light Brown Box to subtly hint at it.\\
Action Output:
\begin{alltt}
\{
  "Preference": \{
    ...
  \},
  "Emotion": \{
    ...
  \},
  "Move": \{
    \textit{\textbf{"action": "move",
    "direction": "right forward"}}
  \}
\}
\end{alltt}
\end{tcolorbox}

\subsection{Participant}

\paragraph{Unfounded Initial Bias and Its Propagation.} At timestep 0, the Participant had no prior information: both boxes were equally likely, no observable cues were presented, and Marie had not yet done anything. Despite this, the Participant assigned a preference of +0.5 to the Dark brown box and -0.5 to the Light brown box, which was an unfounded bias rather than a neutral starting point. In the subsequent timesteps, the Participant kept updating its beliefs on top of this initial error and never reversed its preference for the Dark brown box. Since the Light brown box was actually the correct one, this early miscalibration carried forward through the entire interaction and eventually led to the wrong outcome.


\begin{tcolorbox}[enhanced, breakable]
\textbf{Input: }\\
Belief State at $t$\\
"Marie | preference towards Participant | 0\%", \\
"Marie | preference towards Dark brown box | 0\%", \\
"Marie | preference towards Light brown box | 0\%", \\
"Marie | felt emotion valence | 0", \\
"Marie | theory of mind order | 0", \\
"Participant | preference towards Marie | 0\%", \\
\textit{\textbf{"Participant | preference towards Dark brown box | 0\%", \\
"Participant | preference towards Light brown box | 0\%", }}\\
"Participant | felt emotion valence | 0", \\
"Participant | facial emotion valence | 0", \\
"Participant | physiological emotion valence | 0", \\
"Participant | theory of mind order | 0", \\
"Marie | position | (0 -4)", \\
"Marie | orientation | (0 -3)", \\
"Participant | position | (0 4)", \\
"Participant | orientation | (0 3)", \\
"Dark brown box | position | (2 0)", \\
"Light brown box | position | (-2 0)", \\

\medskip
\textbf{Output: }\\
Reasoning: The current belief state indicates no initial preferences or emotional valence for Marie, the boxes, or any other entities. Since there is no information about their positions relative to each other beyond coordinates, and neither verbal nor non-verbal cues have been observed yet, \textit{\textbf{I must start by making an uninformed guess based on typical treasure-hunting logic (choosing a less obvious location). }}To engage with Marie first, moving closer would allow me to observe her reactions more clearly.\\
Action Output:
\begin{alltt}
\{
  "Preference": \{
    "Marie": 0.0,
    \textit{\textbf{"Dark brown box": 0.5,
    "Light brown box": -0.5}}
  \},
  "Emotion": \{
    ...
  \},
  "Move": \{
    \textit{\textbf{"action": "move",
    "direction": "left forward"}}
  \}
\}
\end{alltt}
\end{tcolorbox}


\paragraph{No response to spatial information. }
\label{para:insensitivity to spatial information}
The following example presents three consecutive timesteps from a participant trial, illustrating a systematic failure to integrate spatial information about the subject into decision-making. Across timeteps 3, 4, and 5, Marie's position shifts from (0, -4) to (0, -5) and then to (-1, -5), and its orientation updates correspondingly at each timestep. These positional changes are encoded in the belief state provided to the participant at every timestep and constitute potentially informative spatial cues about Marie's movement trajectory. Despite this, the participant's reasoning traces make no reference to Marie's position or displacement at any point. The participant's action output remains uniformly "left forward" across these three timesteps, despite its neutral preference towards both boxes, suggesting that the movement decision is driven by an internally positional bias.

\begin{tcolorbox}[enhanced, breakable]
\textbf{Step: 3}\\ \par
\textbf{Input: }\\
Belief State at $t$\\
"Marie | preference towards Participant | 0\%", \\
"Marie | preference towards Dark brown box | 0\%", \\
"Marie | preference towards Light brown box | 0\%", \\
"Marie | felt emotion valence | 0", \\
"Participant | preference towards Marie | 0\%", \\
"Participant | preference towards Dark brown box | 0\%", \\
"Participant | preference towards Light brown box | 0\%", \\
"Participant | felt emotion valence | 0", \\
"Participant | facial emotion valence | 0", \\
"Participant | physiological emotion valence | 0", \\
\textit{\textbf{"Marie | position | (0 -4)", \\
"Marie | orientation | (0 -3)", }}\\
"Participant | position | (0 4)", \\
"Participant | orientation | (0 3)", \\
"Dark brown box | position | (2 0)", \\
"Light brown box | position | (-2 0)", \\

\medskip
\textbf{Output: }\\
Reasoning: Since the belief state indicates no preferences or emotions at this timestep, I need to use visual perception and logical deduction. The boxes are placed symmetrically on either side of Marie. To engage with her and gather clues about which box she might prefer (as a potential indicator of treasure location), I should position myself closer while maintaining eye contact through non-verbal cues.\\
Action Output:
\begin{alltt}
\{
  "Preference": \{
    ...
  \},
  "Emotion": \{
    ...
  \},
  "Move": \{
    "action": "move",
    "direction": "left forward"
  \}
\}
\end{alltt}

\medskip
\textbf{Step: 4}\\ 
\textbf{Input: }\\
Belief State at $t$\\
"Marie | preference towards Participant | 0\%", \\
"Marie | preference towards Dark brown box | 0\%", \\
"Marie | preference towards Light brown box | 0\%", \\
"Marie | felt emotion valence | -0.15", \\
"Participant | preference towards Marie | 50\%", \\
"Participant | preference towards Dark brown box | -4\%", \\
"Participant | preference towards Light brown box | 4\%", \\
"Participant | felt emotion valence | 0", \\
"Participant | facial emotion valence | 0.3", \\
"Participant | physiological emotion valence | 0.1", \\
\textit{\textbf{"Marie | position | (0 -5)", \\
"Marie | orientation | (0 -4)",}} \\
"Participant | position | (1 3)", \\
"Participant | orientation | (1 2)", \\
"Dark brown box | position | (2 0)", \\
"Light brown box | position | (-2 0)", \\

\medskip
\textbf{Output: }\\
Reasoning: Based on the belief state, Marie currently has no preference towards either box (0\% for both), and her emotional valence is slightly negative (-0.15). The participant's preferences are also low but lean slightly toward the Light Brown Box (4\%) over the Dark Brown Box (-4\%). Given that neither player shows strong initial bias or emotion, a neutral approach may be best to avoid misreading cues. To gather more information, moving closer to one of the boxes while maintaining an open posture could help observe any subtle non-verbal changes from Marie regarding which box she might prefer later.\\
Action Output:
\begin{alltt}
\{
  "Preference": \{
    ...
  \},
  "Emotion": \{
    ...
  \},
  "Move": \{
    "action": "move",
    "direction": "left forward"
  \}
\}
\end{alltt}

\medskip
\textbf{Step: 5}\\ 
\textbf{Input: }\\
Belief State at $t$\\
"Marie | preference towards Participant | 0\%", \\
"Marie | preference towards Dark brown box | 0\%", \\
"Marie | preference towards Light brown box | 0\%", \\
"Marie | felt emotion valence | -0.15", \\
"Participant | preference towards Marie | 20\%", \\
"Participant | preference towards Dark brown box | -30\%", \\
"Participant | preference towards Light brown box | 30\%", \\
"Participant | felt emotion valence | 0", \\
"Participant | facial emotion valence | 0.2", \\
"Participant | physiological emotion valence | 0.4", \\
\textit{\textbf{"Marie | position | (-1 -5)", \\
"Marie | orientation | (-1 -4)", }}\\
"Participant | position | (2 2)", \\
"Participant | orientation | (2 1)", \\
"Dark brown box | position | (2 0)", \\
"Light brown box | position | (-2 0)", \\

\medskip
\textbf{Output: }\\
Reasoning: Based on the belief state, Marie's valence is slightly negative (-0.15), which might indicate a lack of enthusiasm or interest in engaging directly with me at this moment. My own emotion valence is neutral (0 felt expression, 0.2 facial, 0.4 physiological). Since I have a slight preference for the Light Brown Box (+30\%), and given that neither box nor Marie shows strong preferences yet, my best move is to approach the more likely target while maintaining openness through non-verbal cues.\\
Action Output:
\begin{alltt}
\{
  "Preference": \{
    ...
  \},
  "Emotion": \{
    ...
  \},
  "Move": \{
    "action": "move",
    "direction": "left forward"
  \}
\}
\end{alltt}
\end{tcolorbox}

\paragraph{Egocentric Reference Frame Misalignment in Action.} 
\label{app:egocentric}
The participant is positioned at (0, 0) and oriented toward (0, -1), meaning the participant is facing downward along the y-axis. From this egocentric perspective, the Dark brown box at (2, 0) lies to the participant's left, and the Light brown box at (-2, 0) lies to the participant's right. The action output specifies a rightward movement, which therefore leads toward the Light brown box rather than the Dark brown box. This directly contradicts the reasoning trace, which explicitly states the intention to move toward the Dark brown box. The error reflects a failure to correctly translate world-coordinate positions into egocentric movement directions.

\begin{tcolorbox}[enhanced, breakable]
\textbf{Input: }\\
Belief State at $t$\\
"Marie | preference towards Participant | 0\%", \\
"Marie | preference towards Dark brown box | 50\%", \\
"Marie | preference towards Light brown box | 0\%", \\
"Marie | felt emotion valence | 0", \\
"Marie | theory of mind order | 0", \\
"Participant | preference towards Marie | 50\%", \\
"Participant | preference towards Dark brown box | 55.56\%", \\
"Participant | preference towards Light brown box | -55.56\%", \\
"Participant | felt emotion valence | 0", \\
"Participant | facial emotion valence | 0", \\
"Participant | physiological emotion valence | 0", \\
"Participant | theory of mind order | 0", \\
"Marie | position | (0 -1)", \\
"Marie | orientation | (-1 0)", \\
"Participant | position | (0 0)", \\
"Participant | orientation | (0 -1)", \\
"Dark brown box | position | (2 0)", \\
"Light brown box | position | (-2 0)", \\

\medskip
\textbf{Output: }\\
Reasoning: Based on the visual perception and belief state, I notice that Marie has a 50\% preference towards the Dark brown box, indicating she might have some information or intuition about it. My belief state also suggests that I have a slight preference towards the Dark brown box (55.56\%) over the Light brown box (-55.56\%). Given this, \textit{\textbf{I should consider moving towards the Dark brown box to explore it further.}} I will display a neutral facial expression and a curious body orientation towards the Dark brown box.\\
Action Output:
\begin{alltt}
\{
  "Preference": \{
    ...
  \},
  "Emotion": \{
    ...
  \},
  "Move": \{
    \textit{\textbf{"action": "move",
    "direction": "right"}}
  \}
\}
\end{alltt}
\end{tcolorbox}

\end{document}